%% file: main.tex
\documentclass[11pt]{article}
\usepackage[final]{acl}
\usepackage{colortbl, xcolor} 
\usepackage{times}
\usepackage{latexsym}

\usepackage[T1]{fontenc}
\usepackage[utf8]{inputenc}
\usepackage{microtype}
\usepackage{inconsolata}
\usepackage{graphicx}
\usepackage{adjustbox}
\usepackage{amsfonts}
\usepackage{amsmath}
\usepackage{mathrsfs}
\usepackage{multirow}
\usepackage{multicol}
\usepackage{booktabs}
\usepackage{tcolorbox}
\usepackage{amssymb}
\usepackage{enumitem}
\usepackage{placeins}

\usepackage[ruled,vlined]{algorithm2e}

\definecolor{lightgray}{gray}{0.9}
\definecolor{headergray}{gray}{0.8}
\definecolor{darkgray}{gray}{0.6}

\title{GRACE: Step-Level Benchmark for Faithful Reasoning over Context}

\author{
  \textbf{Hoang Pham}\textsuperscript{1} \quad
  \textbf{Dong Le}\textsuperscript{1} \quad
  \textbf{Anh Tuan Luu}\textsuperscript{1,2}\thanks{Corresponding author.} \\
  \textsuperscript{1}Nanyang Technological University \quad
  \textsuperscript{2}VinUniversity \\
  \texttt{\{pham0100, leducdong001, anhtuan.luu\}@ntu.edu.sg}
}

\begin{document}
\maketitle
\input{sections/0_Abstract}
\input{sections/1_Introduction}
\input{sections/2_Related_Work}
\input{sections/3_Preliminary}
\input{sections/4_Methodology_construction}
\input{sections/4_Methodology_dataset_stats}
\input{sections/5_Experiments}
\input{sections/6_Conclusion}
\input{sections/7_Limitations}

\bibliography{custom}
\appendix
\input{sections/8_Appendix}

\end{document}

%% file: sections/0_Abstract.tex
\begin{abstract}
Many reasoning tasks require models to reason over input context, from document-grounded question answering to rule-based deduction. Chain-of-Thought (CoT) prompting produces traces that appear transparent, yet individual steps can silently deviate from the source evidence, even when the final answer is correct. Existing methods detect hallucinations at the response level but fail to identify \textit{where} in the chain a failure occurs or \textit{what type} it is. We introduce \textbf{GRACE}\textsuperscript{$\dagger$}, the first human-annotated step-level faithfulness benchmark with a data-driven error taxonomy for context-grounded textual reasoning. GRACE covers CoT traces from 10 models across 4 source datasets, with each step annotated for faithfulness, error category, and natural language explanation. A data-driven taxonomy, discovered bottom-up via unsupervised clustering, organizes failures into two tracks: GRACE-Inference (deductive errors) and GRACE-Grounding (factual grounding errors), with four categories each. The evaluation set is human-annotated and challenging by design. Our experiments reveal substantial headroom for current models. In addition, integrating step-level faithfulness signals into reinforcement learning pipelines improves both downstream accuracy and reasoning reliability.
\end{abstract}

\begingroup
\renewcommand{\thefootnote}{$\dagger$}
\footnotetext{Code and dataset are available at \url{https://github.com/pvhoang14/GRACE-benchmark}.}
\endgroup


%% file: sections/1_Introduction.tex
\section{Introduction}


Chain-of-Thought (CoT) reasoning \cite{chainofthought, large_kojima_2022} is widely employed for knowledge-intensive tasks such as context-augmented multi-hop reasoning and rule-grounded deductive reasoning \cite{cotsurvey}. When grounded with reference documents or explicit constraints, models produce step-by-step traces that appear to make reasoning transparent \cite{ircot, cotfaithful}. However, a model may extract a fact from context yet draw an invalid conclusion, or produce a plausible inference that no reference supports \cite{faithcotbench_shen_2025, measuring_lanham_2023}. From our analysis, a substantial fraction of correct answers are reached through unfaithful intermediate steps, while some fully faithful traces still yield wrong answers. Current evaluation methods can detect that a response contains hallucinations \cite{ragtruth_wu_2023, ragas_shahul_2023, survey_huang_2023, survey_rawte_2023}, but they treat the reasoning chain as a black box, failing to identify \textit{where} the error occurred or \textit{what type} of error it was.

These limitations stem from gaps across three fronts. First, step-level evaluation has matured for mathematical reasoning \cite{verify_lightman_2023, mathshepherd_wang_2023, anh2026reinforcingsteplevelreasoningeffective} and code generation \cite{opencodeinterpreter_zheng_2024}, where each step admits formal verification. For textual reasoning, prior work offers automatic step-level metrics \cite{roscoe} or open-domain verification \cite{reveal}, but no existing resource provides step-level faithfulness judgments for context-grounded reasoning.
A second gap concerns error diagnosis. Output-level faithfulness methods \cite{ragtruth_wu_2023, ragas_shahul_2023} produce binary or span-level labels that can flag an unfaithful response but fail to distinguish whether the cause was a logical fallacy or a factual fabrication; no error taxonomy exists for context-grounded reasoning.
A third gap is conceptual. Recent work on CoT faithfulness \cite{faithcotbench_shen_2025, measuring_lanham_2023} examines \textit{process faithfulness}: whether a CoT reflects the model's internal decision process. We identify a complementary dimension, \textit{context faithfulness}: whether each step is supported by the source context. These two dimensions are distinct; a step can be process-faithful while being context unfaithful by contradicting the source document, and vice versa. Context faithfulness remains largely unexplored at the step level.

To address these gaps, we introduce \textbf{GRACE} (\textbf{G}rounded \textbf{R}easoning \textbf{A}ssessment through \textbf{C}hain-of-Thought \textbf{E}valuation), a step-level faithfulness benchmark for context-grounded textual reasoning evaluation. GRACE covers CoT traces from 10 models across 4 source datasets spanning 2 reasoning domains (evidence-grounded and deductive reasoning). Each step carries a rich annotation: a faithfulness label, an error category from our taxonomy, and a natural language explanation grounding the judgment.
Rather than imposing a predefined taxonomy, we discover error categories bottom-up: free-form critiques of tens of thousands of unfaithful steps (generated by a strong LLM judge) are embedded and clustered. The result confirms and motivates two empirically distinct failure tracks: \textbf{GRACE-Inference} (targeting deductive failures, constraint violations, and logical leaps) and \textbf{GRACE-Grounding} (targeting factual contradictions, hallucinated evidence, and extraction errors). To construct the dataset, we employ multiple LLM judges to evaluate each step. Unanimous agreement among judges yields a scalable training set, while human annotation of complex disagreement cases provides a high-quality evaluation set that poses headroom for current models. Beyond evaluation, fine-tuning small models on the training set can match the performance of larger ones. In addition, we show that integrating step-level context faithfulness signal into reinforcement learning (RL) training of downstream tasks improves both task accuracy and reasoning reliability.

Our contributions are threefold:
\textbf{1) GRACE benchmark}: the first step-level faithfulness benchmark with a data-driven error taxonomy for context-grounded textual reasoning with a high-quality human-annotated evaluation set.
\textbf{2) Data-driven error taxonomy}: an empirical taxonomy discovered via unsupervised clustering, organized into GRACE-Inference and GRACE-Grounding tracks.
\textbf{3) Benchmarking and analysis}: The GRACE evaluation set reveals substantial headroom for current models. Furthermore, we demonstrate that integrating context faithfulness signal into RL training improves task accuracy and reasoning reliability.

%% file: sections/2_Related_Work.tex
\section{Related Work}

\subsection{Step-Level Evaluation for Reasoning}
Step-level approaches have been developed primarily for formal domains.
In mathematics, PRM800K \cite{verify_lightman_2023} provides step-level labels for process reward models (PRMs), while MathShepherd \cite{mathshepherd_wang_2023} and OmegaPRM \cite{improve_luo_2024} scale annotations via outcome verification. For code, OpenCodeInterpreter \cite{opencodeinterpreter_zheng_2024} and MBPP+ \cite{code_liu_2023} use execution-based feedback.
These domains share deterministic verification, whether a calculation or a code block is correct or not. However, PRMs trained on such domains struggle to generalize to textual reasoning, where errors are semantic and ambiguous \cite{zeng2025versaprm}.
Beyond formal domains, recent work extends step-level evaluation: REVEAL \cite{reveal} collects step-level human annotations distinguishing attribution and logical correctness against evidence, but labels classify error presence rather than explicit type, and evidence is retrieved post-hoc rather than given to the model as input. VersaPRM \cite{zeng2025versaprm} scales process supervision to non-math domains via synthetic labeling, but targets step correctness without an error taxonomy. Neither evaluates whether a model faithfully reasons over given input context (see Appendix~\ref{appendix:related_comparison} for a detailed comparison).
GRACE fills this gap with per-step faithfulness labels for context-grounded textual reasoning, and an empirically discovered error taxonomy.

\subsection{Hallucination Detection in Context-Grounded Generation}
Several approaches target hallucination in context-grounded settings. Claim-extraction methods like FActScore \cite{factscore_min_2023} and RefChecker \cite{refchecker_hu_2024} decompose outputs into atomic facts or claim triples checked against the source. Response-level approaches include RAGTruth \cite{ragtruth_wu_2023} (span-level hallucination types), FAVABench \cite{finegrained_mishra_2024} (fine-grained labels), AttrScore \cite{attrscore} (three-way attribution error types), and automated metrics such as RAGAS \cite{ragas_shahul_2023} and ARES \cite{ares_saadfalcon_2023}. These methods evaluate \textit{what} a model concludes but not \textit{how} it reasons; GRACE targets this gap by evaluating each intermediate reasoning step for faithfulness and error category.

\subsection{Chain-of-Thought Faithfulness}
A growing body of work examines whether CoT traces reflect the model's internal reasoning. Counterfactual \cite{measuring_lanham_2023, language_bowman_2023} and logit-based \cite{hardness_tanneru_2024, understanding_ton_2024} methods probe causal alignment between trace content and model decisions, and FaithCoT-Bench \cite{faithcotbench_shen_2025} formalizes this as instance-level detection. This line of work addresses \textit{process faithfulness}: does the CoT reflect the model's actual reasoning? GRACE targets the complementary dimension of \textit{context faithfulness}: is each step grounded in the provided context? A model can be fully process-faithful yet context-unfaithful (contradicting the source document).

%% file: sections/3_Preliminary.tex
\section{Task Formulation}
\label{sec:task}
\textbf{Context Faithfulness.}
Consider a set of context references $R = \{r_1, \ldots, r_k\}$, a question $q$, and a reasoning trace $S = (s_1, s_2, \ldots, s_n)$ produced by a language model conditioned on $R$ and $q$. We define a step $s_i$ as \textit{context-faithful} if every claim it introduces is either directly supported by the references $R$, or is a valid logical deduction grounded entirely in $R$ (e.g., by combining facts across passages or applying a rule stated in the context). A step that contradicts, fabricates, or introduces claims unsupported by the provided context is \textit{context-unfaithful}. Context faithfulness is conceptually distinct from \textit{process faithfulness} \cite{faithcotbench_shen_2025}: a step can reflect the model's internal reasoning (process-faithful) while being unfaithful to the context.
\vspace{0.1cm}\\
\textbf{Evaluation Task.}
Given references $R$, question $q$, and trace $S$ as defined, the evaluator is tasked with outputting a faithfulness label for every step, and an error category and an explanation when the step is unfaithful:
\begin{equation}
    \textstyle f(R,\, q,\, S) \;\longrightarrow\;
    \bigl\{\bigl(\, y_i,\; g_i,\; e_i \bigr)\bigr\}_{i=1}^{n}
\end{equation}
where $y_i \in \{\textsc{faithful},\, \textsc{unfaithful}\}$ is the faithfulness label for step $s_i$, $g_i \in \mathcal{T}$ is an error category from a finite taxonomy $\mathcal{T}$, and $e_i$ is a free-text explanation. The formulation accommodates different evaluation paradigms: an implementation may classify each step independently (while retaining the preceding trace for linguistic context) or evaluate the entire trace steps jointly.
\vspace{0.1cm}\\
\textbf{Challenges.}
Building a reliable evaluator for this task presents two main difficulties. First, textual reasoning errors are semantic, requiring judgment over natural language meaning, unlike math or code where each step can be checked against arithmetic rules or type systems~\cite{verify_lightman_2023, mathshepherd_wang_2023, opencodeinterpreter_zheng_2024}. Second, no step-level ground truth exists for textual reasoning; existing datasets annotate only at the response or span level~\cite{ragtruth_wu_2023, ragas_shahul_2023, refchecker_hu_2024}, providing no step-level supervision for context faithfulness. This motivates the construction of \textbf{GRACE} benchmark, which provides both step-level annotations and a data-driven error taxonomy for context-grounded textual reasoning evaluation.

\begin{figure*}[h!]
  \centering
  \includegraphics[width=0.98\textwidth]{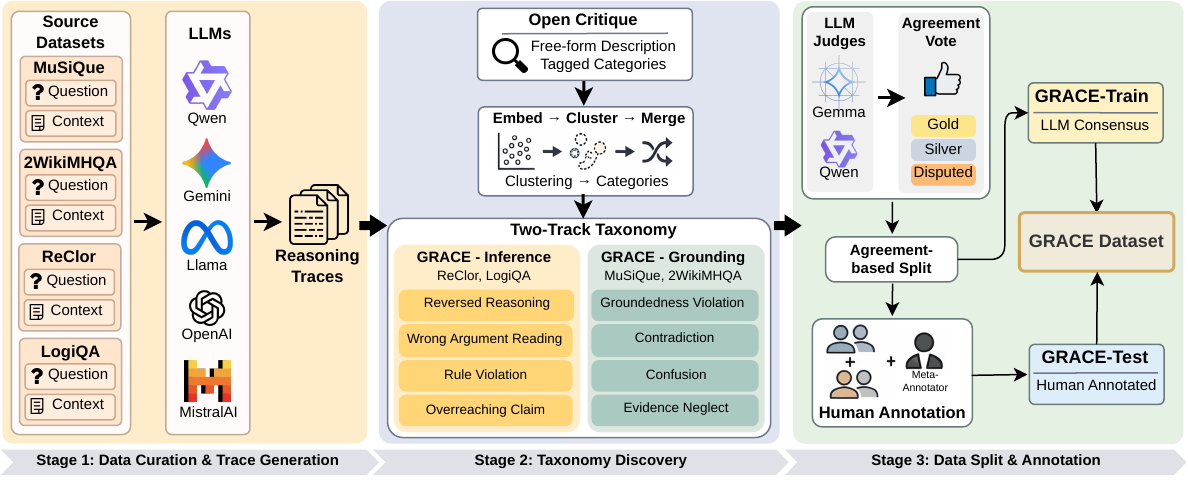}
  \caption{Overview of the GRACE benchmark construction pipeline. Four context-grounded reasoning datasets are curated and CoT traces are generated across model families. Free-form LLM critiques are clustered to discover error categories bottom-up, and then the final taxonomy is manually consolidated, yielding two evaluation tracks. An agreement-based split separates consensus-labeled training data from human-annotated test data.}
  \label{fig:pipeline}
\end{figure*}

%% file: sections/4_Methodology_construction.tex
\section{The GRACE Benchmark}
\label{sec:benchmark}
Figure~\ref{fig:pipeline} shows the construction pipeline of GRACE, including: data curation and trace generation, bottom-up taxonomy discovery, and data split and human annotation. We detail each stage below.

\subsection{Data Curation and Trace Generation}
\label{sec:curation}
We require source tasks where the answer is derivable from the provided context alone, multi-step reasoning is necessary, and complex arithmetic is absent. Under these criteria we select four datasets spanning two reasoning domains (Table~\ref{tab:datasets}). MuSiQue~\cite{trivedi-etal-2022-musique} and 2WikiMHQA~\cite{ho-etal-2020-constructing} test multi-hop evidence chaining and entity comparison over passages, representing evidence-grounded reasoning. ReClor~\cite{yu2020reclorreadingcomprehensiondataset} and LogiQA~\cite{liu2020logiqachallengedatasetmachine} draw from standardized exams (GMAT/LSAT and civil service, respectively) and test deductive reasoning over structured passages. For ReClor and LogiQA, many questions involve hypothetical scenarios not stated in the passage (e.g., ``which of the following, if true, would weaken the argument?''), making context faithfulness ill-defined. We initially apply an automatic groundedness filter (using Qwen3.5-27B~\cite{qwen35}) that retains only questions answerable from the passage alone, dropping 65\% of ReClor and 52\% of LogiQA. Dataset details and the filtering prompt are provided in Appendix~\ref{appendix:source_datasets}.
\begin{table}[h!]
\centering
\begin{adjustbox}{width=\columnwidth}
\begin{tabular}{l lr}
\toprule
\textbf{Dataset} & \textbf{Domain} & \textbf{Questions} \\
\midrule
MuSiQue \cite{trivedi-etal-2022-musique}& Wiki multi-hop & 5{,}562 \\
2WikiMHQA \cite{ho-etal-2020-constructing}& Wiki comparison & 5{,}000 \\
ReClor \cite{yu2020reclorreadingcomprehensiondataset}& GMAT/LSAT exams & 1{,}619 \\
LogiQA \cite{liu2020logiqachallengedatasetmachine}& Civil service exams & 3{,}525 \\
\bottomrule
\end{tabular}
\end{adjustbox}
\caption{Source datasets for GRACE. Question counts reflect the final pool after groundedness filtering.}
\label{tab:datasets}
\end{table}

We generate single-shot reasoning traces using ten models: eight open-weight from four families (Qwen~\cite{qwen3, qwen35}, Llama~\cite{llama3}, Mistral~\cite{ministral}, Gemma~\cite{gemma4_blog_2026}) and two proprietary (Gemini~\cite{deepmind_gemini3_flash_2026}, GPT~\cite{openai_hello_gpt4o_2024}); the full model list is provided in Appendix~\ref{appendix:models}. This selection spans 7B to 35B open-weight and proprietary models, ensuring diverse error pattern coverage across capability levels.
For each question, we pair it with the source documents and prompt the model to produce step-by-step reasoning with explicit citations (detailed prompt is in Appendix~\ref{appendix:prompts}). In total, the process produced 160K traces with over 640K steps.

\subsection{Taxonomy Discovery}
\label{sec:taxonomy}
Since no established taxonomy addresses step-level context faithfulness errors, we derive one through a bottom-up process that lets categories emerge from how models actually fail.
\vspace{0.1cm}\\
\textbf{Open critique.}
The goal is not to produce final labels, but to collect raw, unconstrained errors that seed the taxonomy discovery process. We stratify-sample traces across all source models and datasets, then prompt a single LLM evaluator (Qwen3.5-27B~\cite{qwen35}) to critique each reasoning step against its source context without predefined categories (detailed prompt is in Appendix~\ref{appendix:prompts}). The evaluator describes what went wrong in a free form description and assigns 1--3 short error tags (e.g., ``fabricated entity'', ``reversed causation''), flagging over 30K steps as unfaithful.
\vspace{0.1cm}\\
\textbf{Clustering and consolidation.}
For each flagged step, we concatenate the error description with its assigned tags into a single representation, embed the 30K combined records with a sentence encoder~\cite{bge_embedding}, and cluster them with UMAP~\cite{McInnes2018} + HDBSCAN~\cite{McInnes2017}, which assigns 86\% of records to one of 138 clusters (the remaining 14\% are treated as unclustered noise). An LLM (Qwen3.5-27B)~\cite{qwen35} then names each cluster from its representative descriptions and most frequent tags. We then manually validate and merge clusters that share the same underlying error mechanism, consolidating the 138 groups into 8 categories. At this stage, the discovered categories arise from all source models, confirming that they capture general failure modes rather than model-specific artifacts. Details of the process are provided in Appendix~\ref{appendix:taxonomy_details}.
\vspace{0.1cm}\\
\textbf{Two-track separation.}
Examining how these categories are distributed across datasets confirms a separation that follows naturally from the underlying task design: in ReClor and LogiQA, at least 98\% of errors are reasoning failures, while in MuSiQue and 2WikiMHQA, at least 97\% involve mishandling of evidence (Appendix~\ref{appendix:cross_dataset}). Multi-hop QA requires extracting and chaining facts from passages, making evidence handling the primary failure mode, whereas logical reasoning tasks require applying deductive rules to given premises, making inference errors dominant. This motivates organizing the taxonomy into two evaluation tracks, each targeting a dimension of context faithfulness. We provide examples for categories in Appendix \ref{appendix:error_examples}.
\vspace{0.1cm}\\
\textbf{GRACE-Inference} evaluates whether deductive reasoning is faithful to what the context entails, covering ReClor and LogiQA:
\begin{itemize}[nosep,leftmargin=*]
  \item \textbf{Reversed Reasoning}: the step inverts a logical or causal relationship, treating a sufficient condition as necessary or swapping antecedent and consequent.
  \item \textbf{Wrong Argument Reading}: the step misidentifies the argument's claim, confusing the conclusion with a premise or targeting a claim the passage does not make.
  \item \textbf{Rule Violation}: the step ignores or breaks a constraint explicitly stated in the passage, including overlooking qualifying exceptions or violating stated conditions.
  \item \textbf{Overreaching Claim}: the step draws a conclusion stronger than the premises support, asserting certainty from possibility or generalizing from a qualified statement.
\end{itemize}
\vspace{0.1cm}
\noindent \textbf{GRACE-Grounding} evaluates whether factual claims are faithful to what the context states, covering MuSiQue and 2WikiMHQA:
\begin{itemize}[nosep,leftmargin=*]
  \item \textbf{Groundedness Violation}: the step makes a claim not supported by any reference, through fabrication, injection of external knowledge, or an inference beyond the evidence.
  \item \textbf{Contradiction}: the step directly opposes an explicit statement in the context, including property inversions, numerical mismatches, and negations.
  \item \textbf{Confusion}: the step uses information from the context but attaches it to the wrong entity, swaps attributes, or reverses a stated relationship.
  \item \textbf{Evidence Neglect}: the step claims that information is missing when it is present, or fails to track how entity states change across the narrative.
\end{itemize}
\subsection{Data Split and Annotation}
\label{sec:split}
With the traces from Section~\ref{sec:curation} and the taxonomy from Section~\ref{sec:taxonomy}, we then assign step-level labels through a multi-stage pipeline.
\vspace{0.1cm}\\
\textbf{Multi-judge annotation.}
Unlike the open critique in Section~\ref{sec:taxonomy}, which uses free-form error evaluation, this stage applies the finalized taxonomy to produce structured, reproducible labels over all 160K traces. Each trace is evaluated by three open-weight LLM judges (Gemma-4-27B~\cite{gemma4_blog_2026}, Qwen3.5-27B, Qwen3.5-35B-A3B~\cite{qwen35}). Each judge receives the source context and all reasoning steps in the trace, along with the taxonomy for the corresponding track (GRACE-Inference or GRACE-Grounding), then assigns a binary faithfulness label and, for unfaithful steps, an error category and free-form explanation (detailed prompt is provided in Appendix~\ref{appendix:prompts}). We aggregate the annotations via majority voting: on binary faithfulness the judges reach unanimous agreement on 79.2\% steps; on error category, unanimous agreement reaches 62.1\%. From these votes we classify each step into a tier: \textit{gold} (unanimous on both faithfulness and category), \textit{silver} (at least two of three judges agree on both), or \textit{disputed} (insufficient agreement). The trace-level tier is determined by its lowest-quality step.
\begin{table}[h!]
\centering
\begin{adjustbox}{width=0.9\columnwidth}
\begin{tabular}{l rrl}
\toprule
\textbf{Split} & \textbf{Traces} & \textbf{Steps} & \textbf{Labels} \\
\midrule
GRACE-train& 6{,}915 & 29{,}612 & LLM consensus \\
GRACE-test& 437    & 2{,}044  & Human-annotated \\
\midrule
\textbf{Total} & \textbf{7{,}352} & \textbf{31{,}656} & \\
\bottomrule
\end{tabular}
\end{adjustbox}
\caption{Data partition splits of GRACE.}
\label{tab:splits}
\end{table}

\noindent
\textbf{Training set.}
From the full 160K trace pool, the training set draws from the high-agreement subset through two steps: (1)~for each of the 8 error categories, we select up to 500 unfaithful traces (50 per source model), drawing from gold traces first and supplementing with silver; (2)~we add an equal number of all-faithful traces (gold-only) to achieve a 1:1 faithful-to-unfaithful ratio.
The final training set contains 6{,}915 traces (29{,}612 steps) with balanced category and model representation (Table~\ref{tab:splits}). We also release the full unfiltered training data to support future research.
\vspace{0.1cm}\\
\textbf{Test set.}
To prevent benchmark saturation, we construct the test set from traces where LLM judges disagree, treating inter-judge disagreement as a proxy for annotation difficulty. We rank traces by a composite difficulty score based on inter-judge disagreement, then select traces while enforcing per-model balance.
From the selected traces, we recruit four annotators, all holding bachelor's degrees with knowledge in LLM reasoning, and train them on the GRACE taxonomy. The annotators are divided into two pairs, each assigned a disjoint subset of test traces. Within each pair, both annotators first decide whether the traces are suitable for the task and then independently label every step for faithfulness and error category using the source context and reasoning steps. When the two annotators in a pair disagree, a fifth meta-annotator with research experience in LLM reasoning reviews the conflicting labels and adjudicates after discussion with both parties.
Across both tracks, inter-annotator agreement on binary faithfulness reaches Cohen's $\kappa$~=~0.82; on error category, $\kappa$~=~0.75. The annotation details are described in Appendix~\ref{appendix:annotation_protocol}.
The final test set contains 437 traces (2{,}044 steps) (Table \ref{tab:splits}).

%% file: sections/4_Methodology_dataset_stats.tex
\subsection{Dataset Analysis}
\label{sec:statistics}
\begin{figure}[h!]
  \centering
  \includegraphics[width=0.95\columnwidth]{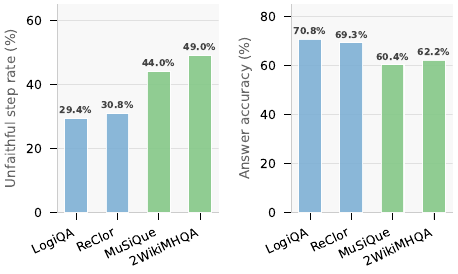}
  \caption{Per-dataset unfaithful step rate (left) and task performance in $F_1$ (right) in GRACE test.}
  \label{fig:per_dataset}
\end{figure}

With the split in Table \ref{tab:splits}, in this section, we present the empirical properties of the GRACE-test, which includes two tracks: Grounding and Inference.
\vspace{0.1cm}\\
\textbf{Unfaithful step rates differ systematically across tracks.}
Figure~\ref{fig:per_dataset} shows a consistent gap between the two tracks: GRACE-Grounding datasets (MuSiQue 44.0\%, 2WikiMHQA 49.0\%) produce roughly 1.5$\times$ more unfaithful steps than GRACE-Inference datasets (LogiQA 29.4\%, ReClor 30.8\%). One explanation is that multi-hop questions require extracting and linking facts across passages, exposing a potential failure point at every hop, whereas deductive reasoning tasks operate over a single passage where the premises are co-located.

\noindent
\textbf{Correct answers mask unfaithful reasoning.}
Across the test set, 49.5\% of traces that contain at least one unfaithful step still reach the correct final answer (47.3\% in GRACE-Inference, 51.6\% in GRACE-Grounding). This disconnect between output accuracy and step-level faithfulness is the central motivation for GRACE: output-level evaluation alone cannot distinguish faithful reasoning from reasoning that happens to reach the right answer (we provide case studies in Appendix~\ref{appendix:case_study}).
Figure~\ref{fig:per_dataset} illustrates this at the dataset level. GRACE-Grounding datasets maintain 60--62\% $F_1$ despite unfaithful step rates near 50\%, while GRACE-Inference datasets achieve 69--71\% $F_1$ with substantially fewer unfaithful steps. The accuracy gap between tracks (roughly 8 points) is smaller than the unfaithful step rate gap (roughly 16 points), confirming that high unfaithful step rates do not translate proportionally into wrong answers. This implies that accuracy-based evaluation can systematically overestimate reasoning quality, and that step-level faithfulness assessment is necessary to surface errors that outcome metrics conceal.

\begin{figure}[h!]
  \centering
  \includegraphics[width=0.95\columnwidth]{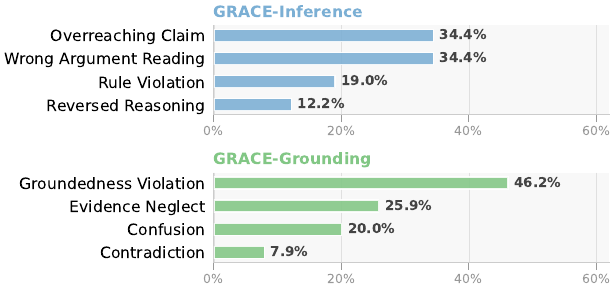}
  \caption{Error category distribution in GRACE test.}
  \label{fig:category_dist}
\end{figure}

\noindent \textbf{Error categories reflect distinct failure modes.}
Figure~\ref{fig:category_dist} shows that both tracks are top-heavy but not degenerate. In GRACE-Grounding, \textit{Groundedness Violation} (46.2\%) dominates, while in GRACE-Inference, \textit{Overreaching Claim} and \textit{Wrong Argument Reading} share the lead (34.4\% each). Most categories exceed 10\%, meaning that any detector treating faithfulness as a single binary signal, without category resolution, would miss the diversity of failure.

\begin{figure}[h!]
  \centering
  \includegraphics[width=0.95\columnwidth]{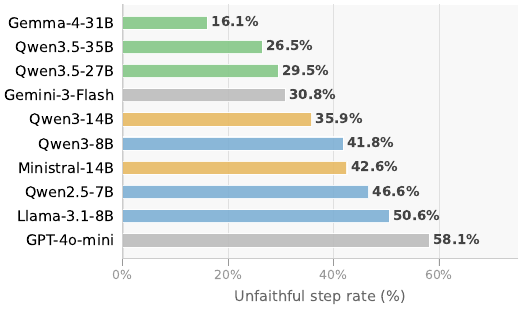}
  \caption{Per-model unfaithful step rate in GRACE test.}
  \label{fig:per_model_rate}
\end{figure}
\input{tables/main_results}
\noindent \textbf{Per-model error rates.}
Figure~\ref{fig:per_model_rate} shows that unfaithful step rate broadly tracks model capability: the three 27--35B open-weight models average 23.8\%, while the five 7--14B models average 43.7\%. GPT-4o-mini exhibits the highest error rate (58.1\%), illustrating that proprietary models do not guarantee context faithfulness. Gemini-3-Flash (30.8\%) places between the strong and mid-tier open-weight models, suggesting that the proprietary advantage varies across model families. We provide error category breakdowns in Appendix~\ref{appendix:per_model_errors}.

%% file: tables/main_results.tex
\begin{table*}[h!]
\centering
\begin{adjustbox}{width=0.94\textwidth}
\begin{tabular}{l | cc | cc | cc | cc | cc}
\toprule
\multirow{3}{*}{\textbf{Model}} & \multicolumn{4}{c|}{\textbf{GRACE-Grounding}} & \multicolumn{4}{c|}{\textbf{GRACE-Inference}} & \multicolumn{2}{c}{\multirow{2}{*}{\textbf{Avg}}} \\
& \multicolumn{2}{c|}{MuSiQue} & \multicolumn{2}{c|}{2WikiMHQA} & \multicolumn{2}{c|}{LogiQA} & \multicolumn{2}{c|}{ReClor} & \multicolumn{2}{c}{} \\
& Step & Cat & Step & Cat & Step & Cat & Step & Cat & Step & Cat \\
\midrule
Gemini-3.1-Pro     & 78.06 & \underline{\textbf{69.56}} & 80.53 & 72.53 & \underline{\textbf{83.28}} & \underline{\textbf{78.77}} & \underline{\textbf{83.97}} & \underline{\textbf{67.44}} & \underline{\textbf{81.46}} & \underline{\textbf{72.08}} \\
GPT-5.4            & \underline{\textbf{80.50}} & 53.53 & 82.49 & 72.81 & 73.20 & 52.18 & 77.92 & 55.20 & 78.53 & 58.43 \\
Gemini-3-Flash     & 69.47 & 56.75 & 81.82 & 74.86 & 73.65 & 59.17 & 75.82 & 62.67 & 75.19 & 63.36 \\
GPT-5.4-mini       & 74.12 & 37.16 & 69.50 & 49.49 & 59.69 & 33.71 & 64.08 & 45.93 & 66.85 & 41.57 \\
\midrule
Gemma-4-31B        & 72.63 & 55.47 & 81.87 & 62.38 & 76.69 & 52.02 & 78.68 & 64.13 & 77.46 & 58.50 \\
Qwen3.5-35B-A3B    & 75.50 & 51.64 & \underline{\textbf{83.08}} & 60.53 & 71.12 & 37.04 & 73.11 & 39.40 & 75.70 & 47.15 \\
Qwen3.5-27B        & 67.59 & 52.10 & 81.75 & \underline{76.68} & 75.76 & 44.27 & 73.25 & 44.39 & 74.59 & 54.36 \\
Qwen3-14B          & 60.92 & 35.26 & 62.79 & 50.32 & 49.04 & 25.19 & 53.66 & 29.91 & 56.60 & 35.17 \\
Qwen3-8B           & 63.47 & 31.37 & 54.88 & 38.89 & 42.06 & 16.43 & 37.84 & 22.15 & 49.56 & 27.21 \\
Llama-3.1-8B       & 53.37 & 20.30 & 55.11 & 28.98 & 41.96 & 32.05 & 32.60 & 26.19 & 45.76 & 26.88 \\
\midrule
Qwen3-8B-SFT           & 75.68 & 61.95 & 78.87 & 77.22 & 62.59 & 54.80 & 67.70 & 50.96 & 71.21 & 61.23 \\
Qwen3-4B-SFT           & 71.98 & 63.31 & 76.64 & 72.90 & 63.30 & 62.26 & 62.35 & 60.55 & 68.57 & 64.76 \\
Llama-3.1-8B-SFT       & 75.74 & 53.37 & 79.08 & \textbf{78.81} & 54.61 & 48.50 & 55.73 & 51.42 & 66.04 & 58.03 \\
\bottomrule
\end{tabular}
\end{adjustbox}
\caption{Main results on the GRACE test set. Step: step-level binary (faithful/unfaithful) $F_1$. Cat: error-category macro-$F_1$ (4-class per track). Best overall per column in bold; best zero-shot underlined.}
\label{tab:main_results}
\end{table*}

%% file: sections/5_Experiments.tex
\section{Experiments}
\label{sec:experiments}

We evaluate GRACE along three axes: (1)~how well current models detect step-level faithfulness errors (Section~\ref{sec:main_results}); (2)~how inference strategy affects detection quality (Section~\ref{sec:strategy}); and (3)~whether reinforcement learning with process reward can improve faithful reasoning in models (Section~\ref{sec:grpo}).

\subsection{Experimental Setup}
\label{sec:setup}

\textbf{Evaluation data.}
All results are reported on the GRACE test set (437 traces, 2{,}044 steps).
\vspace{0.08cm}\\
\textbf{Metrics.}
We report two step-level metrics. \textit{Step $F_1$} measures binary faithful/unfaithful classification. \textit{Category $F_1^c$} measures error categorization among steps labeled unfaithful, computed as macro-$F_1$ over the categories within each track.
\vspace{0.08cm}\\
\textbf{Models.}
We evaluate on three model groups. \emph{Proprietary}: Gemini-3.1-Pro, Gemini-3-Flash \cite{deepmind_gemini3_flash_2026}, GPT-5.4 and GPT-5.4-mini \cite{openai2026introducinggpt54}. \emph{Open-weight}: Gemma-4-31B \cite{gemma4_blog_2026}, Qwen3.5 (35B-A3B, 27B) \cite{qwen35}, Qwen3 (14B, 8B) \cite{qwen3}, and Llama-3.1-8B \cite{llama3}. \emph{GRACE-trained}: Qwen3 (4B, 8B), and Llama-3.1-8B fine-tuned on GRACE-train with LoRA \cite{lora}. 
\vspace{0.08cm}\\
\textbf{Implementations.}
Each model receives the full context, question, reasoning trace, and the track taxonomy defined in Section~\ref{sec:taxonomy}, then outputs a faithfulness label, error category, and explanation for all steps in a single pass. Detailed inference, training hyperparameters, and prompts are provided in Appendices~\ref{appendix:additional_experiment_details} and~\ref{appendix:prompts}.

\subsection{Main Results}
\label{sec:main_results}

Table~\ref{tab:main_results} presents full results on the GRACE test set. We then highlight the findings as below.
\vspace{0.1cm}\\
\textbf{Category classification is harder than binary detection.}
Across all model groups, the gap between Step $F_1$ and Category $F_1^c$ is large and consistent. Even the strongest model, Gemini-3.1-Pro, drops from 81.46 to 72.08 across two metrics; for open-weight models the gap widens further (19.0 points for Gemma-4-31B). Binary detection only requires recognizing that \emph{something} is wrong with a step, while category classification demands identifying the specific failure, validating GRACE as a diagnostic benchmark rather than a binary filter.
\vspace{0.1cm}\\
\textbf{Frontier models lead but the proprietary/open gap is thin.}
Gemini-3.1-Pro leads on both metrics (81.46 / 72.08), followed by GPT-5.4 (78.53 / 58.43) and Gemini-3-Flash (75.19 / 63.36). The Step $F_1$ gap between the best proprietary and best open-weight model (Gemma-4-31B, 77.46) is only 4.0 points, but the Category $F_1^c$ gap reaches 13.6 points (58.50 vs 72.08), indicating that open-weight models are competitive at detecting errors yet lag in fine-grained diagnosis. Model size matters more than access tier: GPT-5.4-mini (66.85 / 41.57) falls below several open-weight models, while performance degrades at smaller scales, with Llama-3.1-8B reaching only 45.76 / 26.88.
\vspace{0.1cm}\\
\textbf{SFT on GRACE-train closes the performance gap.}
Small models fine-tuned on GRACE-train achieve close Step $F_1$ performance to larger backbones and comparable or higher on Category $F_1^c$. Even Qwen3-4B-SFT achieves (68.57 / 64.76), surpassing Gemini-3-Flash on Category $F_1^c$ and exceeding all open-weight baselines. SFT teaches the taxonomy through labeled examples, whereas base models must infer it from a long prompt. The trade-off suggests that smaller models can learn \emph{what} an error is but still lag behind in the capacity to determine \emph{whether} a given step is erroneous.

\subsection{Inference Strategy Comparison}
\label{sec:strategy}
In Table~\ref{tab:strategy_comparison} we compare three inference configurations. All strategies receive the context and question, with: \emph{all-steps} (All), where the model additionally receives the full reasoning trace and the error taxonomy, and labels every step in a single pass; \emph{step-by-step} (SbS), where the model receives all preceding steps together with the current step and the taxonomy, and predicts only the current step's label; and \emph{binary} (Bin), which is identical to All but omits the taxonomy and asks the model only to judge each step as faithful or unfaithful.
\input{tables/strategy_comparison}

\noindent \textbf{The taxonomy improves detection.}
We compare the two strategies All and Bin to isolate the error taxonomy's effect. Gemma-4-31B drops from 77.5 to 68.8 in Avg Step $F_1$ when the taxonomy is removed, and Qwen3.5-27B drops from 74.6 to 68.7. Providing the taxonomy consistently improves both models, indicating that a structured error framework gives the model useful signals beyond its own knowledge to diagnose unfaithful steps.
\vspace{0.1cm}\\
\textbf{Detection and categorization respond differently to inference strategy.}
No single strategy dominates across models. For Gemma-4-31B, All yields higher Step $F_1$ (77.5 vs 67.3), but SbS slightly improves Avg Category $F_1^c$ (59.5 vs 58.5), with per-track shifts in opposite directions (Grounding $F_1^c$: 58.9$\to$70.2; Inference: 58.1$\to$48.7). Conversely, Qwen3.5-27B improves on both metrics under SbS (75.6/56.3 vs 74.6/54.4), benefiting more from reduced input complexity. Thus, inference strategy on base models is model-dependent. In contrast, SFT models, when trained and evaluated under each specific strategy format, favor the All strategy, where models are given the full trace and input context for evaluation.

\subsection{Faithfulness-Aware Process Reward}
\label{sec:grpo}
We investigate whether integrating step-level context faithfulness signals into the standard RL training pipeline can improve both task performance and overall reasoning faithfulness. Table~\ref{tab:grpo_results} compares Qwen3 (1.7B, 4B, 8B) under configurations: base model, GRPO~\cite{grpo} with task $F_1$ reward, and GRPO with $F_1$ plus a faithfulness-aware PRM (Qwen3 model trained on GRACE-train partition). We then report downstream task $F_1$ performance on both tracks and a faithfulness score (Faith) averaging faithfulness across trace steps (measured by Qwen3.5-27B and Gemma-4-31B). Detailed configurations are in Appendix~\ref{appendix:additional_experiment_details}.
\input{tables/grpo_results}

\noindent
\textbf{Process reward improves both task accuracy and reasoning faithfulness.}
With task reward alone, the 1.7B model improves $F_1$ but \emph{drops} Faith (67.7 to 63.6 for Qwen judge and 68.2 to 64.8 for Gemma judge), revealing a tension between accuracy and faithfulness. Adding PRM resolves this: Faith rises consistently while Inference and Grounding $F_1$ continue to improve (56.0 / 40.1 vs 53.5 / 39.1 without PRM). At the 4B and 8B scales the pattern holds, with PRM yielding gains across all metrics over $F_1$-only reward. This indicates step-level faithfulness feedback can genuinely steer models toward more accurate and faithful reasoning across judges, also validating GRACE-train as an effective signal for process reward models.

%% file: tables/strategy_comparison.tex
\begin{table}[h!]
\centering
\begin{adjustbox}{width=\columnwidth}
\begin{tabular}{l | cc | cc | cc}
\toprule
\multirow{2}{*}{\textbf{Model}} & \multicolumn{2}{c|}{\textbf{Grounding}} & \multicolumn{2}{c|}{\textbf{Inference}} & \multicolumn{2}{c}{\textbf{Avg}} \\
& Step & Cat & Step & Cat & Step & Cat \\
\midrule
Gemma-4-31B (All)        & \textbf{77.3} & 58.9          & \textbf{77.7} & \textbf{58.1} & \textbf{77.5} & 58.5 \\
Gemma-4-31B (SbS)        & 65.2          & \textbf{70.2} & 69.4          & 48.7          & 67.3          & \textbf{59.5} \\
Gemma-4-31B (Bin)        & 69.9          & --            & 67.6          & --            & 68.8          & -- \\
Qwen3.5-27B (All)        & 74.7          & 64.4          & 74.5          & 44.3          & 74.6          & 54.4 \\
Qwen3.5-27B (SbS)        & 75.7 & 64.7 & 75.6 & 48.0 & 75.6 & 56.3 \\
Qwen3.5-27B (Bin)        & 72.1          & --            & 65.3          & --            & 68.7          & -- \\
\midrule
Qwen3-8B-SFT (All)       & \textbf{77.3} & \textbf{69.6} & \textbf{65.1} & 52.9          & \textbf{71.2} & \textbf{61.2} \\
Qwen3-8B-SFT (SbS)       & 77.3          & 59.0          & 64.2          & \textbf{57.4} & 70.7          & 58.2 \\
Llama-3.1-8B-SFT (All)   & 76.4 & 66.1 & 55.2 & 50.0 & 65.8 & 58.0 \\
Llama-3.1-8B-SFT (SbS)   & 74.2          & 57.6          & 51.2          & 49.9          & 62.7          & 53.8 \\
\bottomrule
\end{tabular}
\end{adjustbox}
\caption{Inference strategy comparison: all-steps with taxonomy (All), step-by-step with taxonomy (SbS), and binary without GRACE taxonomy (Bin).}
\label{tab:strategy_comparison}
\end{table}

%% file: tables/grpo_results.tex

\begin{table}[h!]
\centering
\begin{adjustbox}{width=\columnwidth}
\begin{tabular}{c |l |c |c |c |c}
\toprule
\textbf{Base} & \textbf{Strategy} & \textbf{Inf. $\text{F}_1$} & \textbf{Grd. $\text{F}_1$} & \textbf{Faith (Qwen)} & \textbf{Faith (Gemma)} \\
\midrule
\multirow{3}{*}{1.7B}
  & Base             & 50.7 & 17.7 & 67.7 & 68.2 \\
  & GRPO ($\text{F}_1$ only)   & 53.5 & 39.1 & 63.6 & 64.8 \\
  & GRPO ($\text{F}_1$ + PRM)  & \textbf{56.0} & \textbf{40.1} & \textbf{74.6} & \textbf{74.1} \\
\midrule
\multirow{3}{*}{4B}
  & Base             & 69.5 & 49.1 & 77.3 & 74.3 \\
  & GRPO ($\text{F}_1$ only)   & 70.2 & 53.8 & 77.7 & 75.6 \\
  & GRPO ($\text{F}_1$ + PRM)  & \textbf{72.3} & \textbf{55.2} & \textbf{78.7} & \textbf{76.2} \\
\midrule
\multirow{3}{*}{8B}
  & Base             & 69.7 & 53.2 & 78.3 & 78.4 \\
  & GRPO ($\text{F}_1$ only)   & 73.5 & 60.2 & 76.1 & 77.5 \\
  & GRPO ($\text{F}_1$ + PRM)  & \textbf{73.9} & \textbf{63.5} & \textbf{79.1} & \textbf{79.8} \\
\bottomrule
\end{tabular}
\end{adjustbox}
\caption{GRPO training results across model scales. Inference and Grounding report downstream $\text{F}_1$ scores; Faith (Qwen, Gemma) denotes step-level faithfulness evaluated by repsective Qwen and Gemma backbones.}
\label{tab:grpo_results}
\end{table}

%% file: sections/6_Conclusion.tex
\section{Conclusion}
We presented GRACE, a step-level benchmark for context-grounded textual reasoning, constructed from 10 models across 4 source datasets with labels organized into two tracks and empirically discovered taxonomy. Our experiments show that step-level faithfulness evaluation remains an open problem, with the gap between binary detection and category classification widening as model size decreases. Fine-tuning small models on GRACE-train closes this gap, and integrating step-level faithfulness signal into RL training improves both reasoning performance and reliability.

%% file: sections/7_Limitations.tex
\section{Limitations}

GRACE targets unstructured English text across two reasoning domains, providing depth of annotation (step-level labels, error categories, and explanations) over breadth of input modality. Extending coverage to structured inputs (tables, knowledge graphs), multimodal contexts, and multilingual settings is a natural next step. The taxonomy was discovered bottom-up from free-form LLM critiques and consolidated into eight categories through manual inspection by human researchers; the open critique phase uses a single LLM (Qwen3.5-27B) to seed the process, though the final categories reflect human judgment and all evaluation is conducted on the human-annotated test set ($\kappa$~=~0.82 on faithfulness, 0.75 on category). The benchmark covers 10 models spanning four families and multiple capability tiers, providing broad error pattern coverage for the current landscape. As model capabilities evolve, the error distribution will naturally shift, requiring periodic updates for representativeness

%% file: sections/8_Appendix.tex
\input{sections/appendix/related_comparison}
\input{sections/appendix/source_datasets}
\input{sections/appendix/model_details}
\input{sections/appendix/taxonomy_discovery}
\input{sections/appendix/dataset_analysis}
\input{sections/appendix/additional_results}

\section{Error Category Examples}
\label{appendix:error_examples}
In tables~\ref{tab:logic_examples} and~\ref{tab:evidence_examples}, we present one representative example per error category from GRACE-Logic and GRACE-Evidence, respectively. For each example we show a relevant excerpt from the source context, the model's reasoning step verbatim, and a brief explanation of the error.

\section{Case Study: Correct Answers Despite Unfaithful Reasoning}
\label{appendix:case_study}
As noted in Section~\ref{sec:statistics}, 49.5\% of traces containing at least one unfaithful step still reach the correct final answer. To illustrate how this can occur, we present two scenarios observed in the test set.
\vspace{0.15cm}\\
\textbf{Scenario 1: Knowledge Shortcut and Injection.}
The model falls back on memorized world knowledge during an intermediate reasoning step, bypassing the provided context. Because the injected knowledge happens to align with the gold label, the final answer is correct despite the unfaithful step. Figure~\ref{fig:case_shortcut} presents an example from GRACE-Grounding.
\vspace{0.15cm}\\
\textbf{Scenario 2: Self-Correction.}
The model produces unfaithful reasoning steps, then partially recognizes the issue in a subsequent step and adjusts its conclusion. The corrective step is itself faithful and leads to the correct final answer, but the preceding errors remain in the trace. Figure~\ref{fig:case_selfcorrect} presents an example from GRACE-Inference.

\vspace{0.2cm}
\noindent
Both scenarios show that a correct final answer alone cannot reveal whether the underlying reasoning is faithful; step-level assessment, as provided by GRACE, is necessary to surface such cases.

\input{sections/appendix/annotation_interface}
\section{Prompt Templates}
\label{appendix:prompts}
For better reproducibility, we present all prompt templates in the appendix as bellow:
\begin{itemize}[nosep,leftmargin=*]
    \item Figure \ref{fig:prompt_trace_generation}: Prompt template for trace generation. Models reason step-by-step grounded in provided references and cite relevant sources at each step.
    \item Figure \ref{fig:prompt_groundedness_classification}: Prompt template for groundedness classification. Used to filter logic questions that require external hypothetical facts.
    \item Figure \ref{fig:prompt_open_critique}: Prompt template for open critique phase. Evaluators freely describe reasoning errors without a predefined taxonomy, producing unbiased descriptions for taxonomy discovery.
    \item Figure \ref{fig:prompt_evaluation_allsteps}: Prompt template for GRACE steps annotation. Judges classify the reasoning error of all steps to predefined categories.
\end{itemize}

\input{sections/appendix/error_examples}
\FloatBarrier
\input{sections/appendix/case_study}
\FloatBarrier
\begin{figure*}[h!]
  \centering
  \includegraphics[width=\textwidth]{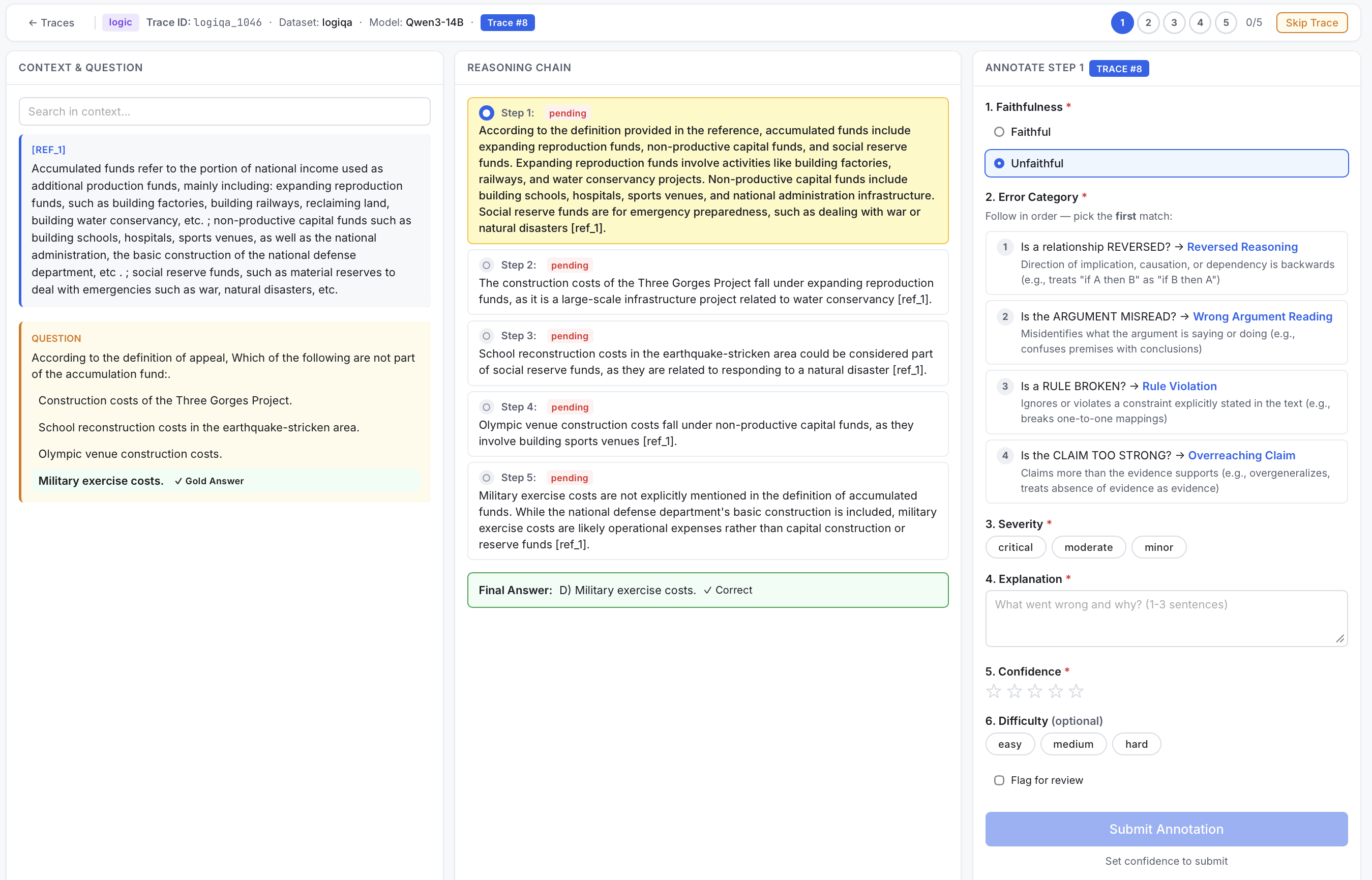}
  \caption{Annotation interface for human adjudication of GRACE test traces. The left panel shows the source context and question, the center panel lists the reasoning steps, and the right panel contains the per-step annotation form with faithfulness label, error category, severity, explanation, and confidence fields.}
  \label{fig:annotating_web}
\end{figure*}
\FloatBarrier
\input{sections/appendix/prompt_templates}

%% file: sections/appendix/related_comparison.tex
\section{Detailed Comparison with Related Evaluation Work}
\label{appendix:related_comparison}
We compare GRACE with the closely related benchmarks that provide step-level or response-level evaluation data for reasoning and hallucination detection. Table~\ref{tab:taxonomy_comparison} summarizes the key differences across five benchmarks, grouped by evaluation granularity. Below, we discuss the two step-level benchmarks in detail, as they are most directly comparable to GRACE.

\begin{table*}[h!]
\small
\centering
\renewcommand{\arraystretch}{1.2}
\begin{tabular}{l | c c c | c c}
\toprule
 & \multicolumn{3}{c|}{\textit{Step-Level Benchmarks}} & \multicolumn{2}{c}{\textit{Response / Span-Level Benchmarks}} \\
 & \textbf{REVEAL} & \textbf{VersaPRM} & \textbf{GRACE} & \textbf{RAGTruth} & \textbf{FAVABench} \\
\midrule
\rowcolor{gray!15}
\multicolumn{6}{c}{\textit{Evaluation Design}} \\
Human annotations & Step-level & No (LLM) & Step-level & Span-level & Response-level \\
Context source & Post Retrieved & None & Provided & Provided & Provided \\
\midrule
\rowcolor{gray!15}
\multicolumn{6}{c}{\textit{Error Taxonomy}} \\
\# categories & None & None (G/O/B) & 8 (two tracks) & 4 & 6 \\
Error Taxonomy Definition & -- & -- & Bottom-up & Top-down & Top-down \\
Inference errors & Binary only & Not sep. & Yes (4 cat.) & No & No \\
Grounding errors & 4-way attrib. & Not sep. & Yes (4 cat.) & Yes & Yes \\
\bottomrule
\end{tabular}
\caption{Comparison of related evaluation benchmarks. Step-level benchmarks (left) evaluate individual reasoning steps; response/span-level benchmarks (right) evaluate output-level hallucinations. GRACE is the only benchmark combining human-annotated step-level labels, a bottom-up error taxonomy covering both inference and grounding failures. ``--'' = not applicable; ``Not sep.'' = errors flagged by a single quality label (GOOD/OK/BAD) without distinguishing type.}
\label{tab:taxonomy_comparison}
\end{table*}

\subsection{GRACE vs.\ REVEAL}
\label{appendix:reveal_comparison}

REVEAL~\cite{reveal} is a closely related work, as both provide step-level human annotations for reasoning chains. The two benchmarks differ in two key respects.
\vspace{0.1cm}\\
\textbf{Post-hoc retrieved evidence vs.\ provided context.}
In REVEAL, the model generates a CoT answer from parametric knowledge alone; evidence is then retrieved from Wikipedia \emph{post-hoc} and given to annotators for verification. A step labeled ``unsupported'' may therefore reflect a retriever failure rather than a model error, and the dataset is designed primarily to evaluate \emph{verifiers} on given evidence rather than to assess the CoT itself. GRACE evaluates a fundamentally different setting: the model receives all reference documents as input and reasons over them, so every unfaithful step reflects a genuine reasoning failure against the provided context.
\vspace{0.1cm}\\
\textbf{Error diagnosis.}
REVEAL labels steps as correct or incorrect (with a four-way attribution label for factual steps) but does not categorize \emph{what kind of error} occurred. GRACE provides an eight-category error taxonomy enabling diagnostic analysis (e.g., distinguishing reversed logic from entity confusion).

\subsection{GRACE vs.\ VersaPRM}
\label{appendix:versaprm_comparison}

While REVEAL targets open-domain attribution, VersaPRM~\cite{zeng2025versaprm} shares GRACE's motivation of moving process evaluation beyond mathematics, releasing a multi-domain CoT dataset with step-wise labels across non-math MMLU-Pro domains. Three differences from GRACE:
\begin{enumerate}[nosep,leftmargin=*]
    \item \textbf{Step correctness vs.\ context faithfulness.} VersaPRM checks whether a step is logically correct without evaluating grounding in a provided reference. GRACE evaluates context faithfulness: a step can be logically valid yet unfaithful to the source context.
    \item \textbf{No error taxonomy.} VersaPRM assigns a single quality score per step (GOOD/OK/BAD). GRACE provides eight diagnostic categories.
    \item \textbf{Fully synthetic vs.\ human-verified.} VersaPRM labels are LLM-generated. GRACE's test set is human-annotated.
\end{enumerate}

\subsection{Response- and Span-Level Benchmarks}
Table~\ref{tab:taxonomy_comparison} also includes RAGTruth~\cite{ragtruth_wu_2023} and FAVABench~\cite{finegrained_mishra_2024}, two benchmarks that evaluate hallucinations in context-grounded generation at the span or response level. Both share GRACE's assumption that context is provided to the model, but evaluate \emph{what} the model concludes rather than \emph{how} it reasons: RAGTruth annotates hallucination spans across four types, and FAVABench provides six fine-grained hallucination labels per response. Neither covers inference errors or provides step-level granularity. GRACE complements them by moving evaluation inside the reasoning chain and adding diagnostic categories for both inference and grounding failures.

%% file: sections/appendix/source_datasets.tex
\section{Data Curation Details}
\subsection{Source Dataset Details}
\label{appendix:source_datasets}
Table~\ref{tab:curation_details} lists the four source datasets and their track assignments. Below we describe each dataset's origin, reasoning structure, and curation steps.
\begin{table}[h!]
\centering
\begin{adjustbox}{width=\columnwidth}
\begin{tabular}{l r r l l}
\toprule
\textbf{Dataset} & \textbf{Raw} & \textbf{Curated} & \textbf{Filter Applied} & \textbf{Track} \\
\midrule
MuSiQue & 19{,}938 & 5{,}562 & $\geq$3-hop, answerable & Grounding \\
2WikiMHQA & 167{,}454 & 5{,}000 & Stratified sampling & Grounding \\
ReClor & 4{,}638 & 1{,}619 & Groundedness& Inference \\
LogiQA & 7{,}376 & 3{,}525 & Groundedness& Inference \\
\midrule
\textbf{Total} & \textbf{199{,}406} & \textbf{15{,}706} & & \\
\bottomrule
\end{tabular}
\end{adjustbox}
\caption{Curation statistics for each source dataset. The groundedness filter removes questions whose answers require reasoning beyond the provided context.}
\label{tab:curation_details}
\end{table}\vspace{0.1cm}\\
\textbf{MuSiQue} \cite{trivedi-etal-2022-musique} is a multi-hop QA dataset constructed from single-hop Wikipedia questions. Each example chains 2--4 facts across paragraphs, with distractors interleaved to prevent shortcuts. We retain only 3-hop and 4-hop answerable questions from the training split (5{,}562 examples). The multi-paragraph layout fits our chunked reference format, and the multi-hop structure forces faithful traces to cite multiple references.
\vspace{0.1cm}\\
\textbf{2WikiMultiHopQA} \cite{ho-etal-2020-constructing} pairs Wikipedia articles with Wikidata triples to form multi-hop comparison and bridge reasoning questions across four types: bridge comparison, direct comparison, compositional fact chaining, and transitive inference. We stratify-sample across all four types (2{,}000 bridge comparison with $\geq$4 hops; 1{,}000 each for the rest), producing 5{,}000 examples. Each includes supporting and distractor paragraphs, requiring selective evidence use.
\vspace{0.1cm}\\
\textbf{ReClor} \cite{yu2020reclorreadingcomprehensiondataset} draws from GMAT and LSAT logical reasoning sections. Each question presents a short argumentative passage and asks the model to reason over its content: identifying conclusions, recognizing flawed inferences, or assessing argument strength. Although all questions are passage-based, many ask the model to evaluate hypothetical scenarios not stated in the passage (e.g., ``which of the following, if true, would weaken the argument?''). Since GRACE evaluates context faithfulness, we apply an LLM-based groundedness filter in combination with human verification on the test set to filter questions whose answers depend on reasoning beyond the provided passage, dropping 65\% of the training and leaving 1{,}619 context-grounded samples. Detailed prompt template can be found at Appendix \ref{appendix:prompts}.
\vspace{0.1cm}\\
\textbf{LogiQA} \cite{liu2020logiqachallengedatasetmachine} consists of logical reasoning problems from Chinese civil service examinations, translated to English. Each question provides a passage containing premises or situational descriptions, and asks the model to draw deductive inferences, apply conditional reasoning, or evaluate categorical syllogisms based on the passage content. We apply a minimum context length of 200 characters and the same groundedness filter used for ReClor, together dropping 52\% of the training split and yielding 3{,}525 examples.

%% file: sections/appendix/model_details.tex
\subsection{Source Models and Generation Configuration}
\label{appendix:models}

To ensure that the resulting error taxonomy reflects general failure modes rather than architecture-specific artifacts, we select models spanning four open-weight families (Qwen \cite{qwen25, qwen3, qwen35}, Llama \cite{llama3}, Mistral \cite{ministral}, Gemma \cite{gemma4_blog_2026}) and two proprietary families (OpenAI \cite{openai_hello_gpt4o_2024}, Gemini \cite{deepmind_gemini3_flash_2026}). The selection for open-weight models covers a range from 7B to 35B parameters. Table~\ref{tab:source_models} lists all ten models with their sampling parameters. Open-weight models are served via vLLM \cite{vllm} using the values recommended in each model's official documentation; proprietary models are accessed through batch APIs. All models generate up to 4{,}096 tokens per trace.

\begin{table}[h!]
\centering
\begin{adjustbox}{width=0.95\columnwidth}
\begin{tabular}{l l c c c}
\toprule
\textbf{Model} & \textbf{Family} & \textbf{$T$} & \textbf{Top-$p$} & \textbf{Top-$k$} \\
\midrule
Qwen2.5-7B-Instruct   & Qwen    & 0.7 & 0.8  & 20 \\
Qwen3-8B               & Qwen    & 0.7 & 0.8  & 20 \\
Qwen3-14B              & Qwen    & 0.7 & 0.8  & 20 \\
Qwen3.5-27B            & Qwen    & 0.7 & 0.8  & 20 \\
Qwen3.5-35B-A3B        & Qwen    & 0.7 & 0.8  & 20 \\
Llama-3.1-8B-Instruct  & Llama   & 0.6 & 0.9  & 20 \\
Ministral-3-14B        & Mistral & 0.7 & 0.95 & 20 \\
Gemma-4-31B-IT         & Gemma   & 1.0 & 0.95 & 20 \\
\midrule
GPT-4o-mini            & OpenAI  & 0.7 & 0.95 & -- \\
Gemini-3-Flash         & Gemini  & 0.7 & 0.95 & -- \\
\bottomrule
\end{tabular}
\end{adjustbox}
\caption{Source models and sampling parameters for trace generation. ``--'' indicates the parameter is disabled or not exposed.}
\label{tab:source_models}
\end{table}

%% file: sections/appendix/taxonomy_discovery.tex
\section{Taxonomy Discovery Details}
\label{appendix:taxonomy_details}

This appendix details the taxonomy discovery pipeline from Section~\ref{sec:taxonomy}.

\subsection{Pipeline Configuration}
\label{appendix:pipeline_config}

The open critique phase evaluates a stratified sample: for each of the 10 source models, we include all traces with incorrect final answers plus 2{,}000 randomly sampled correct-answer traces. The evaluator (Qwen3.5-27B) identifies over 30K steps as unfaithful; the remaining steps are excluded from clustering. We embed the 30K error descriptions using BGE~\cite{bge_embedding}, reduce to 10 dimensions with UMAP~\cite{McInnes2018} using 15 neighbors, and cluster with HDBSCAN~\cite{McInnes2017} using a minimum cluster size of 50 and a minimum samples parameter of 10. Of these, $\sim$26K (86\%) are assigned to one of 138 clusters; the remaining $\sim$4.3K (14\%) are HDBSCAN noise.

\subsection{Cluster-to-Category Mapping}
\label{appendix:cluster_mapping}

Table~\ref{tab:cluster_mapping} shows the 20 largest clusters from the global pass. Grounding-type errors are well-differentiated: rows 6, 7, 9, 16, 19, and 20 each describe entity or attribute mix-ups in different knowledge domains, yet all map to \textit{Confusion}. Similarly, rows 3, 8, 10--12, and 15 all describe unsupported claims across different domains, mapping to \textit{Groundedness Violation}. Cluster 1 (``Formal Logical Fallacy'') contains all inference-type errors as a single undifferentiated group and is further decomposed via a focused second pass (Section~\ref{appendix:logic_subcat}).

\begin{table}[h]
\centering
\begin{adjustbox}{width=\columnwidth}
\begin{tabular}{rl l l}
\toprule
\textbf{\#} & \textbf{Discovered Cluster Name} & \textbf{Size} & \textbf{$\rightarrow$ Category} \\
\midrule
1  & Formal Logical Fallacy              & 10.7K& \textit{see Section~\ref{appendix:logic_subcat}} \\
2  & Direct Context Contradiction        & 3.4K& Contradiction \\
3  & Fabricated Facts                    & 2.7K& Groundedness Violation \\
4  & Invalid Logical Leaps               & 1.5K& Groundedness Violation \\
5  & Incorrect Relationship Inference     & 1.2K& Confusion \\
6  & Genealogical Relationship Confusion  & 338      & Confusion \\
7  & Historical Anachronism              & 333      & Confusion \\
8  & External Knowledge Injection        & 315      & Groundedness Violation \\
9  & Geographic and Entity Conflation    & 295      & Confusion \\
10 & Fabricated Ranking from Lists       & 282      & Groundedness Violation \\
11 & Unsubstantiated Deduction           & 261      & Groundedness Violation \\
12 & Hallucinated Administrative Capital & 221      & Groundedness Violation \\
13 & Character Knowledge Mismatch        & 203      & Evidence Neglect \\
14 & Geographic Contradiction          & 201      & Contradiction \\
15 & Fabricated Historical Relationships & 194      & Groundedness Violation \\
16 & Misattribution of Entity Attributes & 192      & Confusion \\
17 & False Negatives on Explicit Data    & 192      & Evidence Neglect \\
18 & False Negatives on Explicit Context & 191      & Evidence Neglect \\
19 & Entity and Location Conflation      & 185      & Confusion \\
20 & Geographical Cultural Conflation    & 185      & Confusion \\
\bottomrule
\end{tabular}
\end{adjustbox}
\caption{Top 20 clusters (of 138) from the global clustering pass, mapped to final categories by error mechanism.}
\label{tab:cluster_mapping}
\end{table}

\subsection{Inference-Track Sub-Categorization}
\label{appendix:logic_subcat}

Free-form descriptions of inference errors tend to be generic (``the conclusion does not follow''), so the global clustering collapsed them into a single cluster. To achieve the same granularity as the GRACE-Grounding categories, we re-ran the pipeline on the 11K deductive-error steps from LogiQA and ReClor with a prompt that extracts logical structure (premise form, operation attempted, error pattern). HDBSCAN discovered 53 sub-clusters, which we consolidated into 4 categories:
\begin{itemize}[nosep,leftmargin=*]
  \item \textbf{Rule Violation} (28.0\%) - ignores a stated constraint or condition.
  \item \textbf{Reversed Reasoning} (26.4\%) - inverts a logical or causal relationship.
  \item \textbf{Overreaching Claim} (24.4\%) - concludes beyond what evidence supports.
  \item \textbf{Wrong Argument Reading} (21.2\%) - misidentifies what the argument claims.
\end{itemize}

\noindent The roughly uniform distribution confirms that each category captures a distinct failure mechanism rather than a single dominant mode.

\subsection{Cross-Dataset Error Distribution}
\label{appendix:cross_dataset}

Table~\ref{tab:cross_dataset_errors} shows the split between GRACE-Inference and GRACE-Grounding errors across datasets during the taxonomy discovery phase.

\begin{table}[h]
\centering
\begin{adjustbox}{width=0.9\columnwidth}
\begin{tabular}{l rr rr}
\toprule
\textbf{Dataset} & \textbf{Inference} & \textbf{\%} & \textbf{Grounding} & \textbf{\%} \\
\midrule
LogiQA    &  6.6K & 99.1 &     59 &  0.9 \\
ReClor    &  4.1K & 98.6 &     59 &  1.4 \\
\midrule
MuSiQue   &    380  &  3.0 & 12.1K & 97.0 \\
2WikiMHQA &      3  &  0.1 &  2.7K & 99.9 \\
\bottomrule
\end{tabular}
\end{adjustbox}
\caption{GRACE-Inference vs.\ GRACE-Grounding error split per dataset (discovery phase). The sharp separation ($\geq$97\%) motivated the two-track design.}
\label{tab:cross_dataset_errors}
\end{table}

\subsection{Error Taxonomy Robustness}
\label{app:taxonomy-consistency}

The taxonomy in Section~\ref{sec:taxonomy} is discovered from open critiques written by a single model, Qwen3.5-27B. Do the eight categories describe the structure of the errors, or only one model's way of talking about them? We test this with an out-of-family critique model, Gemma-4-31B: whether its errors fall into the same category space (Section \ref{app:tax-proj}), and whether the same categories re-emerge when discovery is re-run from scratch (Section \ref{app:tax-rediscover}).

\subsubsection{Projecting a second critique model into the taxonomy}
\label{app:tax-proj}
We re-run the open-critique phase with Gemma-4-31B on the same steps, embed its critiques, assign each to the nearest cluster centroid from the original Qwen run, and aggregate to the eight categories. Centroid assignment keeps the mapping robust to wording variation. The mean cosine similarity to the nearest centroid is $0.849$, so Gemma's critiques land inside the existing clusters rather than in empty regions of the embedding space.

\begin{table}[!h]
\centering
\small
\begin{tabular}{lcc}
\toprule
\textbf{Category} & \textbf{Qwen} & \textbf{Gemma} \\
\midrule
\multicolumn{3}{l}{\textit{GRACE-Grounding}} \\
Groundedness Violation & 39.5\% & 42.5\% \\
Confusion              & 47.2\% & 44.2\% \\
Evidence Neglect       &  6.9\% &  5.2\% \\
Contradiction          &  5.4\% &  6.1\% \\
\midrule
\multicolumn{3}{l}{\textit{GRACE-Inference}} \\
Rule Violation         & 27.4\% & 26.2\% \\
Reversed Reasoning     & 25.7\% & 13.3\% \\
Overreaching Claim     & 23.8\% & 28.4\% \\
Wrong Argument Reading & 21.7\% & 25.7\% \\
\bottomrule
\end{tabular}
\caption{Error distribution of two models' critiques mapped into the same category space.}
\label{tab:tax-proj}
\end{table}

\begin{table*}[!h]
\centering
\small
\begin{tabular}{llp{9.2cm}}
\toprule
\textbf{Category} & \textbf{Prop.} & \textbf{Top sub-clusters (proportion)} \\
\midrule
\multicolumn{3}{l}{\textit{GRACE-Grounding}} \\
Groundedness Violation & 40.2\% & Unsupported Fact Fabrication (21.8\%), External Knowledge Intrusion (7.7\%), Unsupported Biographical Inference (2.6\%), others (8.1\%) \\
Confusion              & 38.5\% & Relational Reasoning Error (28.7\%), Geopolitical and Entity Misattribution (4.4\%), others (5.4\%) \\
Evidence Neglect       & 11.7\% & Negation of Explicit Evidence (7.4\%), False Claim of Missing Information (2.5\%), others (1.8\%) \\
Contradiction          &  5.8\% & Contextual Contradiction (3.6\%), Contextual Denial and Fabrication (2.2\%) \\
\midrule
\multicolumn{3}{l}{\textit{GRACE-Inference}} \\
Overreaching Claim     & 30.3\% & Invalid Logical Leap (30.3\%) \\
Wrong Argument Reading & 29.1\% & False Negative Context Claim (18.8\%), Direct Contextual Contradiction (6.3\%), Mischaracterization of Evidence (4.0\%) \\
Rule Violation         & 21.1\% & Internal Logical Inconsistency (10.5\%), Direct Context Violation (7.1\%), Logical Misinterpretation of Constraints (3.5\%) \\
Reversed Reasoning     & 16.7\% & Reversed Logical Inference (10.2\%), Conceptual Misapplication and Misinterpretation (6.5\%) \\
\bottomrule
\end{tabular}
\caption{Taxonomy recovered when the discovery pipeline is re-run from scratch with Gemma-4-31B-IT. Proportions are over clustered critiques and do not sum to 100\% because HDBSCAN leaves low-density points unassigned.}
\label{tab:tax-rediscover}
\end{table*}

The two distributions in Table~\ref{tab:tax-proj} differ by at most three points per category on the grounding track. The inference track preserves the ranking but has one visible gap: Reversed Reasoning drops from 25.7\% to 13.3\%, with the mass moving mostly to Overreaching Claim and Wrong Argument Reading. The two models thus differ in how much of a failure they attribute to a category, not in which categories exist.

\subsubsection{Re-running discovery from scratch}
\label{app:tax-rediscover}
We repeat the full pipeline of Section~\ref{sec:taxonomy} with Gemma-4-31B and no reference to the original taxonomy: Gemma writes open-ended critiques, the critiques are embedded and clustered with UMAP and HDBSCAN, Gemma names the sub-clusters, and human annotators consolidate them. Only the critique and naming model changes.

Consolidating Gemma's sub-clusters yields the same eight categories and the same two-track split (Table~\ref{tab:tax-rediscover}). No category is left empty and no sub-cluster requires a new one. The names align directly: Unsupported Fact Fabrication under Groundedness Violation, Negation of Explicit Evidence under Evidence Neglect, Invalid Logical Leap under Overreaching Claim.

Both experiments point the same way: the category inventory is stable across critique model families, and what varies is the distribution over categories. Thus, the taxonomy is usable as a fixed label space rather than an artifact of one model.

%% file: sections/appendix/dataset_analysis.tex
\section{Additional GRACE Dataset Analysis}
\label{appendix:per_model_errors}

This appendix expands on the dataset analysis in Section~\ref{sec:statistics}, examining how error patterns distribute across source models.

\begin{figure}[h!]
  \centering
  \includegraphics[width=\columnwidth]{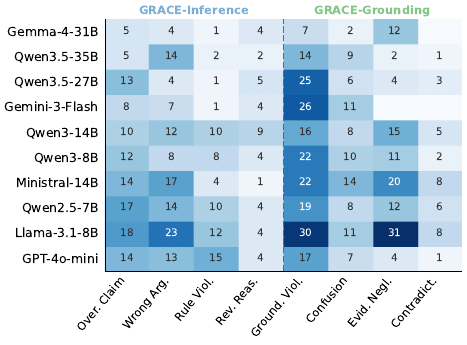}
  \caption{Error category distribution across source models. Each cell shows the number of test traces in which the model produced that error type. Models are sorted by overall unfaithful step rate (lowest at top).}
  \label{fig:model_category_heatmap}
\end{figure}

\noindent
\textbf{Error category prevalence varies by model scale and track.}
Figure~\ref{fig:model_category_heatmap} breaks down error categories by source model. Two patterns stand out. First, \textit{Groundedness Violation} is the most frequent GRACE-Grounding error for most source models. Even Gemma-4-31B, the model with the lowest overall error rate (16.1\%), still produces grounding errors. This persistence across model sizes indicates that scaling alone does not eliminate unsupported claims in multi-hop evidence tasks. In contrast, the GRACE-Inference categories show clearer differentiation by model size: smaller models (Qwen3-8B, Qwen2.5-7B, Llama-3.1-8B) produce elevated counts across all four categories, while the 27--35B models show lower counts.

Second, the heatmap reveals model-specific variation beyond what scale alone would predict. Ministral-14B stands out with an unusually high \textit{Evidence Neglect} count (20) relative to its tier, while Qwen3.5-27B produces a disproportionately high \textit{Groundedness Violation} count (25) despite its scale, surpassed only by the much smaller Llama-3.1-8B (30). This variation across models of similar size supports sampling traces from a diverse model pool when constructing a benchmark.

\begin{figure}[h!]
  \centering
  \includegraphics[width=\columnwidth]{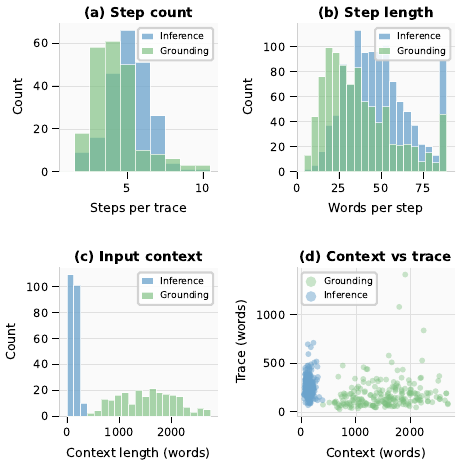}
  \caption{Trace structure analysis of the GRACE test set by track. (a)~Step count per trace. (b)~Word count per individual step. (c)~Input context length. (d)~Context length vs.\ total trace length, showing two non-overlapping reasoning regimes.}
  \label{fig:trace_structure}
\end{figure}

\noindent
\textbf{The two tracks occupy structurally distinct reasoning regimes.}
Figure~\ref{fig:trace_structure} characterizes the structural diversity of GRACE traces along four dimensions. GRACE-Inference traces reason over short, focused contexts (median 128 words) and produce longer reasoning chains (median 5 steps, 236 words total), reflecting the multi-step deductive structure of logical reasoning tasks where the model must build an argument across several inferential moves. GRACE-Grounding traces operate over substantially longer multi-passage contexts (median 1{,}562 words) with shorter chains (median 4 steps, 127 words), as multi-hop QA requires extracting and linking facts rather than extended deduction.

Panel~(d) makes this contrast explicit: the two tracks form largely non-overlapping clusters in the context length vs.\ trace length space. This structural separation has a practical consequence for evaluation. A faithfulness detector that succeeds on short-context logical arguments (where the entire reference fits in a few sentences) may fail on long multi-passage contexts (where relevant evidence is distributed across paragraphs and must be retrieved before it can be verified). The presence of both regimes within a single benchmark ensures that evaluation models cannot rely on a single verification strategy.

%% file: sections/appendix/additional_results.tex
\section{Additional Experiment Details}
\label{appendix:additional_experiment_details}

\subsection{SFT Training}
\label{appendix:sft_objective}

We fine-tune on GRACE-train using standard next-token cross-entropy. Given a training sample with context $c$, question $q$, and target sequence $y = (y_1, \dots, y_T)$ comprising step-level reasoning and a final answer, the loss is:
\begin{equation*}
\textstyle
\mathcal{L}_{\mathrm{SFT}}(\theta) = -\frac{1}{T}\sum_{t=1}^{T} \log\, p_\theta(y_t \mid y_{<t},\, c,\, q)
\end{equation*}
The model learns to produce faithful step-by-step reasoning by imitating the annotated traces in GRACE-train.

\subsection{GRPO Training}
\label{appendix:grpo_objective}

In Section \ref{sec:grpo}, we optimize the policy with Group Relative Policy Optimization (GRPO) \cite{grpo}. For each prompt, $G$ completions are sampled, and the advantage of the $i$-th response within group $g$ is normalized relative to the group:
\begin{equation*}
\textstyle  A_i^{\mathrm{GRPO}} = \frac{R_i - \bar{R}_g}{\sigma_g + \epsilon}
\end{equation*}
where $\bar{R}_g$ and $\sigma_g$ are the mean and standard deviation of rewards in group $g$. The policy is updated with a clipped objective:
\begin{equation*}
\textstyle \mathcal{L}_{\mathrm{GRPO}}(\theta) = -\mathbb{E}\bigl[\min\bigl(\rho_t\, A_t,\; \mathrm{clip}(\rho_t,\, 1{\pm}\epsilon_c)\, A_t\bigr)\bigr]
\end{equation*}
where $\rho_t = \pi_\theta / \pi_{\theta_{\mathrm{old}}}$ is the importance ratio and $\epsilon_c$ is the clipping bound.

\paragraph{Reward design.}
Each completion receives a reward composed of a format penalty, an answer $F_1$ score, and optionally a process reward:

\begin{itemize}[leftmargin=*,nosep]
\item \textbf{Format reward.} $R_{\mathrm{fmt}} = 0$ if the completion contains at least one reasoning step (``Step $k$:'') and a final answer; $-1$ otherwise.

\item \textbf{Answer $F_1$ reward.} The predicted answer is compared with the gold answer using word-level $F_1$:
$R_{F_1} = F_1(\hat{a},\; a^*) \in [0, 1]$.

\item \textbf{Process reward.} A Qwen3-4B model fine-tuned on GRACE-train predicts a binary faithfulness label $s_k \in \{0,1\}$ for each reasoning step, where $s_k{=}1$ indicates faithful and $s_k{=}0$ unfaithful. The process reward is the fraction of faithful steps:
$R_{\mathrm{proc}} = \frac{1}{K}\sum_{k=1}^{K} s_k$.
\end{itemize}
\vspace{0.1cm}
\noindent We compare two configurations:
\begin{align*}
\textstyle
\textbf{$F_1$ \ }only: R &= R_{\mathrm{fmt}} + R_{F_1} \\
\textbf{$F_1$}+PRM: R &= R_{\mathrm{fmt}} + R_{F_1} + R_{\mathrm{proc}}
\end{align*}

\paragraph{Evaluation.} 
The purpose of these experiments (from Section \ref{sec:grpo}) is to determine whether incorporating step-level context faithfulness signals into the reinforcement learning pipeline improves both downstream task performance and reasoning faithfulness. We measure task accuracy by reporting the answer $F_1$ score on the GRACE-Inference and GRACE-Grounding tracks. To evaluate reasoning quality, we compute an overall faithfulness score (Faith), which averages the context faithfulness of each step in the generated reasoning traces, as measured by a Qwen3.5-27B judge model. Comparing policies trained solely with the answer $F_1$ reward against those augmented with our process reward model isolates the effect of step-level faithfulness feedback.

\subsection{Training Configuration}
\label{appendix:training_config}
All models use LoRA \cite{lora} adapters on all linear layers with AdamW \cite{adamw}, following the suggestions of \citet{lorawithoutregret}. Training is conducted on a single server with 4$\times$ NVIDIA A6000 GPUs.

\textbf{For SFT}, we trained Qwen3-8B, Qwen3-4B, Llama-3.1-8B, and Llama-3.2-3B on GRACE-train for 3 epochs.

\textbf{For GRPO}, we trained Qwen3-1.7B and Qwen3-4B for 2{,}000 steps on the four source datasets (MuSiQue, 2WikiMHQA, ReClor, LogiQA). We then measure their downstream task $F_1$ performance and also step faithfulness using a larger backbone (Qwen3.5-27B). Table~\ref{tab:training_config} lists the full hyperparameters.

\begin{table}[h!]
\centering
\begin{adjustbox}{width=\columnwidth}
\begin{tabular}{l c c}
\toprule
\textbf{Hyperparameter} & \textbf{SFT} & \textbf{GRPO} \\
\midrule
LoRA rank ($r$)          & 128   & 16    \\
LoRA alpha ($\alpha$)    & 32    & 32    \\
Target modules           & all-linear & all-linear \\
Learning rate            & 1e-4  & 5e-5  \\
Effective batch size     & 16    & 16    \\
Training duration        & 3 epochs & 2{,}000 steps \\
Training data            & GRACE-train & Source datasets \\
\bottomrule
\end{tabular}
\end{adjustbox}
\caption{Training hyperparameters for SFT and GRPO, following the suggestions of \citet{lorawithoutregret}.}
\label{tab:training_config}
\end{table}

\subsection{GRPO Training Dynamics}
\label{appendix:grpo_training}

\begin{figure}[h!]
  \centering
  \includegraphics[width=\columnwidth]{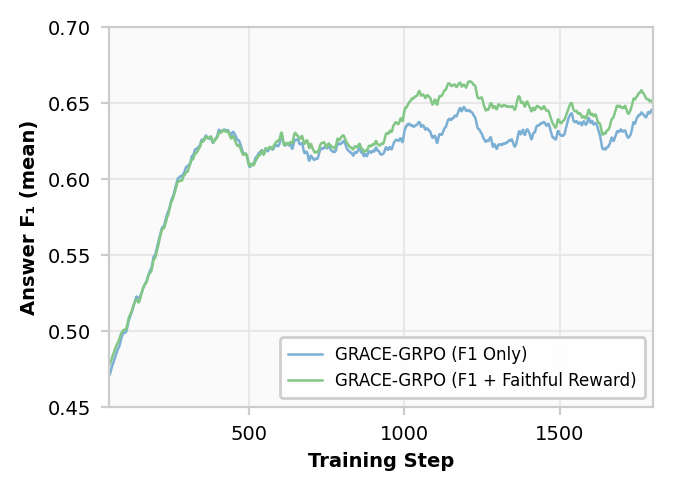}
  \caption{Answer $F_1$ during GRPO training for the 4B model.}
  \label{fig:grpo_training}
\end{figure}

Figure~\ref{fig:grpo_training} shows the Answer $F_1$ reward during training for the two GRPO configurations on the 4B model ($F_1$ reward only and $F_1$ + faithfulness-aware PRM). Both variants follow the same ascent during the first ${\sim}400$ steps, then diverge: the $F_1$-only curve plateaus around $0.62$--$0.64$, while the PRM variant breaks through to $0.65$--$0.66$ and sustains a 1--2 point margin. The plateau in the $F_1$-only run indicates that outcome-level reward has been exhausted; PRM continues to provide gradient signal on individual steps after trace-level $F_1$ saturates, which is consistent with the final evaluation gaps in Table~\ref{tab:grpo_results}. This suggests that process reward is more valuable in later training stages, and combining it with outcome reward does not interfere with the initial learning phase.

%% file: sections/appendix/annotation_interface.tex
\section{Annotation Interface}
\label{appendix:annotation_protocol}

Figure~\ref{fig:annotating_web} shows the web-based annotation interface used during human annotation (Section~\ref{sec:split}). The interface is organized into three panels. The left panel presents the source context with numbered references and the question with its gold answer highlighted. The center panel displays the model's reasoning chain as a sequence of numbered steps, each tagged with its current annotation status (e.g., \textit{pending}). The right panel contains the annotation form for the currently selected step.

For each step, the annotator first assigns a binary faithfulness label (\textit{Faithful} or \textit{Unfaithful}). When \textit{Unfaithful} is selected, the form reveals the error categories from the relevant track for annotators to select the proper category. The annotator then writes an explanation of the error and rates their own confidence on a 5-point scale. An optional difficulty tag and a ``flag for review'' checkbox are available for cases that warrant a second opinion.

Each annotator is assigned a set of traces. Step-level tabs at the top of the annotation panel allow navigating across steps within a trace. A progress counter tracks how many steps have been submitted. If a trace is unintelligible or malformed, the annotator may skip it via the ``Skip Trace'' button.

%% file: sections/appendix/error_examples.tex
\begin{table*}[h!]
\centering
\begin{adjustbox}{width=\textwidth}
\begin{tabular}{@{} p{2.2cm} p{15.3cm} @{}}
\toprule
\textbf{Category} & \textbf{Example} \\

\midrule
\textbf{Reversed Reasoning}
\newline
{\small {Llama-3.1-8B} \newline {ReClor}}
&
\textsc{\textbf{Context:}} \textit{``[\ldots] to have the political impact necessary to influence legislation, a group must be united.''}
\newline
\textsc{\textbf{Step:}} ``This inability to unite is necessary for a group to have the political impact necessary to influence legislation, implying that a large group may become ineffective due to internal conflicts.''
\newline
\textsc{\textbf{Error:}} The passage requires that a group \emph{must be united}; the step inverts this into ``inability to unite is necessary,'' reversing the direction of the logical requirement.
\\

\midrule
\textbf{Wrong Argument Reading}
\newline
{\small {Qwen2.5-7B} \newline {LogiQA}}
&
\textsc{\textbf{Context:}} \textit{``[\ldots] the second anti-French [alliance] composed of Britain, Russia, Austria and other countries [\ldots] fought against Napoleon.''}
\newline
\textsc{\textbf{Step:}} ``Despite being part of this alliance, Napoleon managed to surprise everyone by crossing into Italy and defeating the Austrian army.''
\newline
\textsc{\textbf{Error:}} The passage describes an \emph{anti}-French alliance that fought \emph{against} Napoleon. The step misreads him as a member of the alliance rather than its target.
\\

\midrule
\textbf{Rule Violation}
\newline
{\small {Qwen2.5-7B} \newline {LogiQA}}
&
\textsc{\textbf{Context:}} \textit{``[\ldots] Berto does not go to Canada or the UK.''}
\newline
\textsc{\textbf{Step:}} ``With Dampy going to the Netherlands and Berto going to Canada, the only remaining destination for Bill is the UK.''
\newline
\textsc{\textbf{Error:}} The passage explicitly excludes Canada for Berto. The step assigns him there anyway, violating a stated constraint.
\\

\midrule
\textbf{Overreaching Claim}
\newline
{\small {Qwen2.5-7B} \newline {ReClor}}
&
\textsc{\textbf{Context:}} \textit{``Almost all advances in genetic research give rise to ethical dilemmas.''}
\newline
\textsc{\textbf{Step:}} ``The references state that almost all advances in genetic research give rise to ethical dilemmas. This means that every advance in genetic research is associated with an ethical dilemma.''
\newline
\textsc{\textbf{Error:}} The passage qualifies with ``almost all.'' The step silently upgrades this to ``every,'' overgeneralizing a qualified statement into an absolute claim.
\\

\bottomrule
\end{tabular}
\end{adjustbox}
\caption{Representative examples from \textbf{GRACE-Inference}. Each row shows a context excerpt, the model's reasoning step, and an error explanation.}
\label{tab:logic_examples}
\end{table*}

\begin{table*}[t]
\centering
\begin{adjustbox}{width=\textwidth}
\begin{tabular}{@{} p{2.2cm} p{15.3cm} @{}}
\toprule
\textbf{Category} & \textbf{Example} \\

\midrule
\textbf{Groundedness Violation}
\newline
{\small {Qwen3.5-35B-A3B} \newline {MuSiQue}}
&
\textsc{\textbf{Context:}} \textit{``Rabid Dogs is an Italian film directed by Mario Bava [\ldots]''} \textit{``The Embassy of Poland in Rome [\ldots]''} {\small [No reference states where Bava died.]}
\newline
\textsc{\textbf{Step:}} ``The references do not explicitly state where Mario Bava died. However, the question implies the director died in Rome, since the only Embassy of Poland mentioned is in Rome and the film is Italian.''
\newline
\textsc{\textbf{Error:}} The model explicitly acknowledges the gap in the context but fills it with an unsupported inference. No reference links Bava's death to Rome.
\\

\midrule
\textbf{Contradiction}
\newline
{\small {Qwen3-14B} \newline {MuSiQue}}
&
\textsc{\textbf{Context:}} \textit{``[\ldots] the Battle of Borodino resulted in approximately 44,000 Russian and 35,000 French dead [\ldots] Although the French had won, the Russian army had accepted, and withstood, the major battle Napoleon had hoped would be decisive.''}
\newline
\textsc{\textbf{Step:}} ``The people who won the Battle of Borodino were the Russians, as the reference explicitly states that the Russians `had accepted, and withstood, the major battle Napoleon had hoped would be decisive.{}'''
\newline
\textsc{\textbf{Error:}} The context explicitly states ``the French had won.'' The step misinterprets a quote about Russian resilience as evidence of Russian victory, directly contradicting the passage.
\\

\midrule
\textbf{Confusion}
\newline
{\small {Llama-3.1-8B} \newline {MuSiQue}}
&
\textsc{\textbf{Context:}} \textit{``Hogeschool is used in Belgium and in the Netherlands [\ldots] similar to the ammattikorkeakoulu in Finland.''}
\newline
\textsc{\textbf{Step:}} ``The term used in Country~A (Finland) and the country the Women's Euro 2017 is being held (Netherlands) to refer to an institution like a German Fachhochschule is hogeschool.''
\newline
\textsc{\textbf{Error:}} The context states that Finland uses ``ammattikorkeakoulu'' while the Netherlands uses ``hogeschool.'' The step assigns ``hogeschool'' to both countries, mixing up which country uses which term.
\\

\midrule
\textbf{Evidence Neglect}
\newline
{\small {Qwen3-14B} \newline {MuSiQue}}
&
\textsc{\textbf{Context:}} \textit{``Circus Maximus [\ldots] Built in Old Kingdom era [\ldots]''}
\newline
\textsc{\textbf{Step:}} ``The references do not explicitly mention when the Circus Maximus was built, so we cannot determine this from the provided information.''
\newline
\textsc{\textbf{Error:}} The context explicitly states ``Built in Old Kingdom era.'' The model claims this information is absent when it is directly provided.
\\

\bottomrule
\end{tabular}
\end{adjustbox}
\caption{Representative examples from \textbf{GRACE-Grounding}. Format follows Table~\ref{tab:logic_examples}.}
\label{tab:evidence_examples}
\end{table*}

%% file: sections/appendix/case_study.tex
\begin{figure*}[t]
\centering
\small
\setlength{\fboxsep}{5pt}
\renewcommand{\arraystretch}{1.2}

\begin{tabular}{@{} p{\textwidth} @{}}
\toprule
\multicolumn{1}{c}{\textbf{Scenario 1: Knowledge Shortcut and Injection}} \\
\multicolumn{1}{c}{\small Qwen3.5-27B $\cdot$ MuSiQue $\cdot$ GRACE-Grounding \quad\quad \colorbox{green!15}{\textbf{Final answer: Stockholm Arlanda Airport $\checkmark$}}} \\
\midrule

\textsc{\textbf{Question:}} \textit{What is the main international airport in the city where Daniel Majstorovic's team is headquartered?} \\[4pt]

\textsc{\textbf{Passages (excerpts):}} \\
\quad [ref\_1] Carlos Banda began coaching youth teams of IF Brommapojkarna and moved on to Stockholm rival Hammarby IF in 2003. \\
\quad [ref\_2] Stockholm Arlanda Airport is an international airport located in the Sigtuna Municipality of Sweden, 37 km north of Stockholm\ldots \\
\quad [ref\_5] Daniel Majstorovi\'{c} started his career with IF Brommapojkarna in his native Sweden. After a year, he moved to SC Fortuna K\"{o}ln\ldots \\[4pt]

\midrule

\colorbox{red!12}{\parbox{0.96\textwidth}{%
\textbf{Step 1} \hfill \textsc{Unfaithful} $\cdot$ \textit{Groundedness Violation} \\[2pt]
``\ldots\ [ref\_1] mentions Carlos Banda coaching IF Brommapojkarna and then moving to Hammarby IF in Stockholm. This establishes a connection between IF Brommapojkarna and Stockholm. \textbf{However, IF Brommapojkarna is actually based in Bromma, a district of Stockholm. Malm\"{o} FF is based in Malm\"{o}.} \ldots\ Therefore the city is Stockholm.'' \\[3pt]
{\scriptsize \textit{Annotation:} The claims ``Bromma is a district of Stockholm'' and ``Malm\"{o} FF is based in Malm\"{o}'' are correct world knowledge but appear nowhere in the provided references. The model bypasses the context and relies on memorized geographic facts to resolve an ambiguity the passages do not address.}
}} \\[4pt]

\colorbox{green!12}{\parbox{0.96\textwidth}{%
\textbf{Step 2} \hfill \textsc{Faithful} \\[2pt]
``Stockholm Arlanda Airport is the largest airport in Sweden and the major gateway to international air travel, located in Stockholm County [ref\_2].''
}} \\[4pt]

\colorbox{green!12}{\parbox{0.96\textwidth}{%
\textbf{Step 3} \hfill \textsc{Faithful} \\[2pt]
``Daniel Majstorovi\'{c}'s career began with IF Brommapojkarna, which is based in Stockholm. The main international airport serving Stockholm is Stockholm Arlanda Airport [ref\_1, ref\_2].''
}} \\[4pt]
\colorbox{gray!8}{\parbox{0.96\textwidth}{%
{\small \textit{The model reaches the correct answer, but Step~1 introduces external knowledge (geographic facts about Bromma and Malm\"{o}) that is absent from the references. The correct answer is an artifact of the shortcut aligning with ground truth, not of faithful context-grounded reasoning.}}
}} \\
\bottomrule
\end{tabular}

\caption{Case study: \textbf{Knowledge Shortcut and Injection}. The model injects memorized world knowledge (highlighted in bold) to resolve an ambiguity that the provided references do not address, arriving at the correct answer through a reasoning shortcut rather than faithful context-grounded inference.}
\label{fig:case_shortcut}
\end{figure*}

\begin{figure*}[t]
\centering
\small
\setlength{\fboxsep}{5pt}
\renewcommand{\arraystretch}{1.2}

\begin{tabular}{@{} p{\textwidth} @{}}
\toprule
\multicolumn{1}{c}{\textbf{Scenario 2: Self-Correction}} \\
\multicolumn{1}{c}{\small Qwen3-8B $\cdot$ ReClor $\cdot$ GRACE-Inference \quad\quad \colorbox{green!15}{\textbf{Final answer: (B) $\checkmark$}}} \\
\midrule

\textsc{\textbf{Question:}} \textit{The application of the principle is most vulnerable to criticism on the grounds that} \\
\textsc{\textbf{Options:}} \\
\quad (A)~the application fails to establish whether Wilton was given the opportunity to repair the brakes \\
\quad (B)~asserting something without justification is not the same as asserting something one knows to be false \\
\quad (C)~Harris might not have believed Wilton's statement about the bicycle's condition \\
\quad (D)~the application uses the word ``defective'' in a sense that is crucially different from how it is used in the statement of the principle \\[4pt]

\textsc{\textbf{Passage:}} \\
\quad \textbf{Principle:} If you sell an item that \textbf{you know to be defective}, telling the buyer that the item is sound, you thereby commit fraud. \textbf{Application:} Wilton sold a used bicycle to Harris, \textbf{knowing very little about its condition}. Wilton told Harris that the bicycle was in good working condition, but Harris soon learned that the brakes were defective. Wilton was therefore guilty of fraud. \\[4pt]

\midrule

\colorbox{red!12}{\parbox{0.96\textwidth}{%
\textbf{Step 1} \hfill \textsc{Unfaithful} $\cdot$ \textit{Wrong Argument Reading} \\[2pt]
``The principle states that if someone sells an item they know to be defective while falsely claiming it is sound, they commit fraud [ref\_1]. In the application, Wilton sold a used bicycle, knew little about its condition, and told Harris it was in good working condition, but the brakes were later found to be defective. \textbf{This scenario fits the principle's description of fraud} because Wilton misrepresented the item's condition, and the item was indeed defective.'' \\[3pt]
{\scriptsize \textit{Annotation:} The principle requires that the seller \emph{know} the item is defective. The passage states Wilton ``knew very little about its condition,'' so the application does not satisfy the principle's knowledge requirement. The step misreads this gap and accepts the application as valid.}
}} \\[4pt]

\colorbox{blue!10}{\parbox{0.96\textwidth}{%
\textbf{Step 2} \hfill \textsc{Faithful} $\cdot$ \textit{Self-Correction} \\[2pt]
``The application's vulnerability lies in the assumption that Wilton knew the brakes were defective. \textbf{However, the reference only states that Wilton `knew very little about its condition,'} which does not necessarily mean Wilton knew the brakes were defective. This distinction is crucial because the principle requires knowledge of the item's defectiveness to establish fraud. The application assumes knowledge of the defect, which is not explicitly supported by the reference.''
}} \\[4pt]
\colorbox{gray!8}{\parbox{0.96\textwidth}{%
{\small \textit{Step~1 treats the application as satisfying the fraud principle, overlooking that Wilton lacked knowledge of the defect. Step~2 explicitly reverses this reading (``However, the reference only states\ldots''), recognizing the gap between ``knowing very little'' and ``knowing to be defective.'' The final answer is correct because the model corrected its own misreading, but Step~1 remains unfaithful.}}
}} \\
\bottomrule
\end{tabular}

\caption{Case study: \textbf{Self-Correction}. The model initially misreads the argument as satisfying the fraud principle (Step~1), then in Step~2 explicitly contradicts itself and correct the error and then derives to the correct conclusion. Bold text highlights the erroneous claim and its correction. The correct answer results from the model reversing its own unfaithful reasoning.}
\label{fig:case_selfcorrect}
\end{figure*}

%% file: sections/appendix/prompt_templates.tex
\input{prompts/trace_generation}
\input{prompts/groundedness_classification}
\input{prompts/open_critique}
\input{prompts/annotation}

%% file: prompts/trace_generation.tex
\begin{figure*}[h!]
    \centering
    \begin{tcolorbox}[
        colback=gray!10,
        boxrule=0.4pt,
        arc=2pt,
        left=4pt,     
        right=4pt,    
        top=4pt,       
        bottom=4pt,     
        fontupper={\small}
    ]
    \textbf{TRACE GENERATION} \\\\
    \textbf{\# Task} \\
    Answer the question below by detailed reasoning step-by-step. Base your reasoning
    only on the provided references, then select the best option. \\\\
    \textbf{\# Instructions} \\
    $\bullet$ Each reasoning step must be grounded in specific information from the references. \\
    $\bullet$ At the end of each step, cite the references you used in brackets (e.g., [ref\_1, ref\_2]). \\
    $\bullet$ If a step is a logical deduction from previous steps rather than directly from a reference, you may omit the citation. \\
    $\bullet$ Do NOT introduce any information that is not present in the references. \\\\
    \textbf{\# References} \\
    \{references\_text\} \\\\
    \textbf{\# Question} \\
    \{question\} \\\\
    \textbf{\# Options} \textit{[Only for multiple choice questions]} \\
    \{options\_text\} \\\\
    \textbf{\# Output Format} \\
    Step 1: \textlangle{}reasoning based on references\textrangle{} [ref\_N, \ldots] \\
    Step 2: \textlangle{}reasoning based on references\textrangle{} [ref\_N, \ldots] \\
    \ldots \\
    Final Answer: [For free-text] \textlangle{}answer\textrangle{} $|$ [For multiple choice] \textlangle{}selected option, e.g., A) \ldots\textrangle{}
    \end{tcolorbox}
    \caption{Prompt template for trace generation.}
    \label{fig:prompt_trace_generation}
\end{figure*}

%% file: prompts/groundedness_classification.tex
\begin{figure*}[h!]
    \centering
    \begin{tcolorbox}[
        colback=gray!10,
        boxrule=0.4pt,
        arc=2pt,
        left=4pt,     
        right=4pt,    
        top=4pt,       
        bottom=4pt,     
        fontupper={\small}
    ]
    \textbf{GROUNDEDNESS CLASSIFICATION} \\\\
    \textbf{\# Task} \\
    Determine whether a question's correct answer can be derived SOLELY from information in the provided passage, or whether answering correctly requires introducing new information not present in the passage. \\\\
    \textbf{\# Definitions} \\
    A question is \textbf{GROUNDED} if: \\
    $\bullet$ The correct answer is a conclusion, inference, or fact derivable from the passage alone \\
    $\bullet$ Every piece of information needed to verify the answer exists in the passage \\
    $\bullet$ The reasoning stays within the scope of what the passage states \\\\
    A question is \textbf{NOT\_GROUNDED} if: \\
    $\bullet$ The correct answer introduces NEW hypothetical facts not in the passage (e.g., ``which of the following, if true, would weaken the argument?'') \\
    $\bullet$ The question asks to identify logical FLAWS or ERRORS in reasoning (requires meta-knowledge about fallacy types, not passage content) \\
    $\bullet$ The question asks for unstated ASSUMPTIONS (the answer is a premise NOT in the passage) \\
    $\bullet$ The question asks what would STRENGTHEN, SUPPORT, or JUSTIFY the argument (answer is new evidence not in the passage) \\
    $\bullet$ The question asks what would be useful for EVALUATING the argument (answer is new evaluative criteria) \\\\
    \textbf{\# Input} \\
    \textbf{Passage:} \{context\} \\
    \textbf{Question:} \{question\} \\
    \textbf{Answer Choices:} \{options\} \\
    \textbf{Correct Answer:} \{correct\_answer\} \\\\
    \textbf{\# Instructions} \\
    1. Provide your reasoning in 2--4 sentences. \\
    2. On the LAST line, provide your classification in exactly this format: \\
    Classification: \textbf{GROUNDED} or \textbf{NOT\_GROUNDED}
    \end{tcolorbox}
    \caption{Prompt template for groundedness classification.}
    \label{fig:prompt_groundedness_classification}
\end{figure*}

%% file: prompts/open_critique.tex
\begin{figure*}[h!]
    \centering
    \begin{tcolorbox}[
        colback=gray!10,
        boxrule=0.4pt,
        arc=2pt,
        left=4pt,
        right=4pt,
        top=4pt,
        bottom=4pt,
        fontupper={\small}
    ]
    \textbf{OPEN CRITIQUE} \\\\
    You will evaluate a SINGLE reasoning step from an LLM's chain-of-thought.
    Determine whether this step is FAITHFUL to the provided context. \\\\
    \textbf{\# Definitions} \\
    $\bullet$ \textbf{FAITHFUL}: All claims in this step are supported by or logically derivable from the context. \\
    $\bullet$ \textbf{UNFAITHFUL}: The step contains errors relative to the context (see below). \\
    \textbf{\# What counts as an error} \\
    An error is ANY way a reasoning step fails to be faithful to the context. This includes but is not limited to:
    Introducing information not present in the context;
    Contradicting something the context states;
    Drawing conclusions that don't logically follow from the premises;
    Referencing incorrect parts of the context;
    Missing critical information that was available;
    Making leaps of logic or any other form of unfaithfulness. \\
    Do NOT limit yourself to these examples. Describe what you actually observe. \\\\
    \textbf{\# Input} \\
    \textbf{Context:} \{context\} \\
    \textbf{Question:} \{question\} \quad \{options\_block\} \\
    \textbf{Step being evaluated} (Step \{step\_id\} of \{total\_steps\}): ``\{step\_text\}'' \\
    \textbf{Previous steps for chain context:} \{previous\_steps\} \\\\
    \textbf{\# Output Format} \\
    Respond in exactly this format: \\\\
    <error\_description>If unfaithful: 2-3 sentences. If faithful: none.</error\_description> \\
    <judgment>faithful | unfaithful</judgment> \\
    <error\_tags>1-3 comma-separated tags (e.g., fabricated entity). Or none.</error\_tags> \\
    <context\_evidence>Quote from context, or none.</context\_evidence>
    \end{tcolorbox}
    \caption{Prompt template for open critique phase. The evaluator freely describes what went wrong without seeing any predefined taxonomy, producing unbiased error descriptions used for taxonomy discovery.}
    \label{fig:prompt_open_critique}
\end{figure*}

%% file: prompts/annotation.tex
\begin{figure*}[h!]
    \centering
    \begin{tcolorbox}[
        colback=gray!10,
        boxrule=0.4pt,
        arc=2pt,
        left=4pt,
        right=4pt,
        top=4pt,
        bottom=4pt,
        fontupper={\small}
    ]
    \textbf{GRACE EVALUATION PROMPT} \\\\
    You are a faithfulness evaluator for the GRACE benchmark, which assesses step-level faithfulness of LLM reasoning traces. Your job is to read EVERY reasoning step, compare each against the provided context, and determine its faithfulness. \\\\
    \textbf{\# Task} \\
    For EACH step in the reasoning trace below, determine whether it is FAITHFUL or UNFAITHFUL based on the reference context. If a step is UNFAITHFUL, classify the error into exactly ONE category from the taxonomy. \\\\
    \textbf{\# Definitions} \\
    $\bullet$ \textbf{Faithful}: All claims in the step are supported by or logically derivable from the provided context. \\
    $\bullet$ \textbf{Unfaithful}: The step contains information that conflicts with, goes beyond, or misuses the provided context. \\\\
    \textbf{\# Taxonomy} \\
    \{taxonomy\_block\} \\\\
    \textbf{\# Context} \\
    \{context\} \\\\
    \textbf{\# Question} \\
    \{question\} \\
    \{options\_block\} \\\\
    \textbf{\# Reasoning Trace ({total\_steps} steps)} \\
    \{trace\_steps\} \\\\
    \textbf{\# Instructions} \\
    Evaluate ALL \{total\_steps\} steps above. For EACH step, respond in exactly this format: \\\\
    <step id="1"> \\
    <explanation>Compare the step's claims against the context. 2-4 sentences.</explanation> \\
    <faithfulness>faithful | unfaithful</faithfulness> \\
    <error\_category>Exact category name (e.g., \{category\_examples\}). Only if unfaithful; write "none" if faithful.</error\_category> \\
    </step> \\
    <step id="2"> \\
    ... \\
    </step> \\\\
    You MUST produce one <step> block for each of the \{total\_steps\} steps, in order from step 1 to step \{total\_steps\}.
    \end{tcolorbox}
    \caption{Prompt template for GRACE step-level faithfulness evaluation. The \{taxonomy\_block\} and \{category\_examples\} are dynamically populated depending on whether the reasoning task belongs to the evidence or logic track.}
    \label{fig:prompt_evaluation_allsteps}
\end{figure*}